\documentclass[10pt,twocolumn,letterpaper]{article}

\usepackage{cvpr}              
\usepackage{booktabs}
\usepackage{multirow}
\usepackage{tabularx}
\usepackage{caption}
\usepackage{booktabs,multirow,makecell}
\usepackage{amssymb} 
\usepackage{stfloats}  

\usepackage{float}

\def\@fnsymbol#1{\ensuremath{\ifcase#1\or *\or \dagger\or \ddagger\or
   \mathsection\or \mathparagraph\or \|\or **\or \dagger\dagger
   \or \ddagger\ddagger \else\@ctrerr\fi}}

\newcommand{\ssymbol}[1]{^{\@fnsymbol{#1}}}

\newcommand{\cmark}{$\checkmark$}
\newcommand{\xmark}{$\times$}
\newcolumntype{Y}{>{\centering\arraybackslash}X}  
\definecolor{cvprblue}{rgb}{0.21,0.49,0.74}
\usepackage[pagebackref,breaklinks,colorlinks,allcolors=cvprblue]{hyperref}

\usepackage{lmodern} 
\usepackage[T1]{fontenc} 

\title{PReD: An LLM-based Foundation Multimodal Model for Electromagnetic Perception, Recognition, and Decision}

\author{
\small
Zehua Han\textsuperscript{3,*,\#}\hspace{0.6em}
Jing Xiao\textsuperscript{10,*}\hspace{0.6em}
Yiqi Duan\textsuperscript{11,*}\hspace{0.6em}
Mengyu Xiang\textsuperscript{13,*}\hspace{0.6em}
Yuheng Ji\textsuperscript{12}\hspace{0.6em}
Xiaolong Zheng\textsuperscript{12} \\
\small
Chenghanyu Zhang\textsuperscript{5}\hspace{0.6em}
Zhendong She\textsuperscript{4}\hspace{0.6em}
Junyu Shen\textsuperscript{1}\hspace{0.6em}
Dingwei Tan\textsuperscript{9}\hspace{0.6em}
Shichu Sun\textsuperscript{8}\hspace{0.6em}
Cong Zhou\textsuperscript{6} \\
\small
Mingxuan Liu\textsuperscript{7}\hspace{0.6em}
Fengxiang Wang\textsuperscript{2,\dag}\hspace{0.6em}
Jinping Sun\textsuperscript{3,\dag}\hspace{0.6em}
Yangang Sun\textsuperscript{1,\#,\dag} \\[0.8em]
\scriptsize
\textsuperscript{1}Tsinghua University \quad
\textsuperscript{2}National University of Defense Technology \quad
\textsuperscript{3}Beihang University \\
\scriptsize
\textsuperscript{4}Tianjin University \quad
\textsuperscript{5}Beijing University of Posts and Telecommunications \quad
\textsuperscript{6}Peking University \\
\scriptsize
\textsuperscript{7}Institute of Microelectronics, Chinese Academy of Sciences \quad
\textsuperscript{8}Institute of Software, Chinese Academy of Sciences \\
\scriptsize
\textsuperscript{9}The Hong Kong University of Science and Technology \quad
\textsuperscript{10}Civil Aviation University of China \\
\scriptsize
\textsuperscript{11}Beijing Institute of Technology \quad
\textsuperscript{12}Institute of Automation, Chinese Academy of Sciences \quad
\textsuperscript{13}University of Chinese Academy of Sciences \\[0.5em]
\scriptsize * These authors contributed equally to this work. \quad
\# These authors are the Project Leaders. \quad
$\dag$ These authors are the Corresponding Authors.
}

\begin{document}


 \twocolumn[{%
 \maketitle
 \begin{figure}[H]
      \hsize=\textwidth 
     \centering
     \includegraphics[width=1\textwidth]{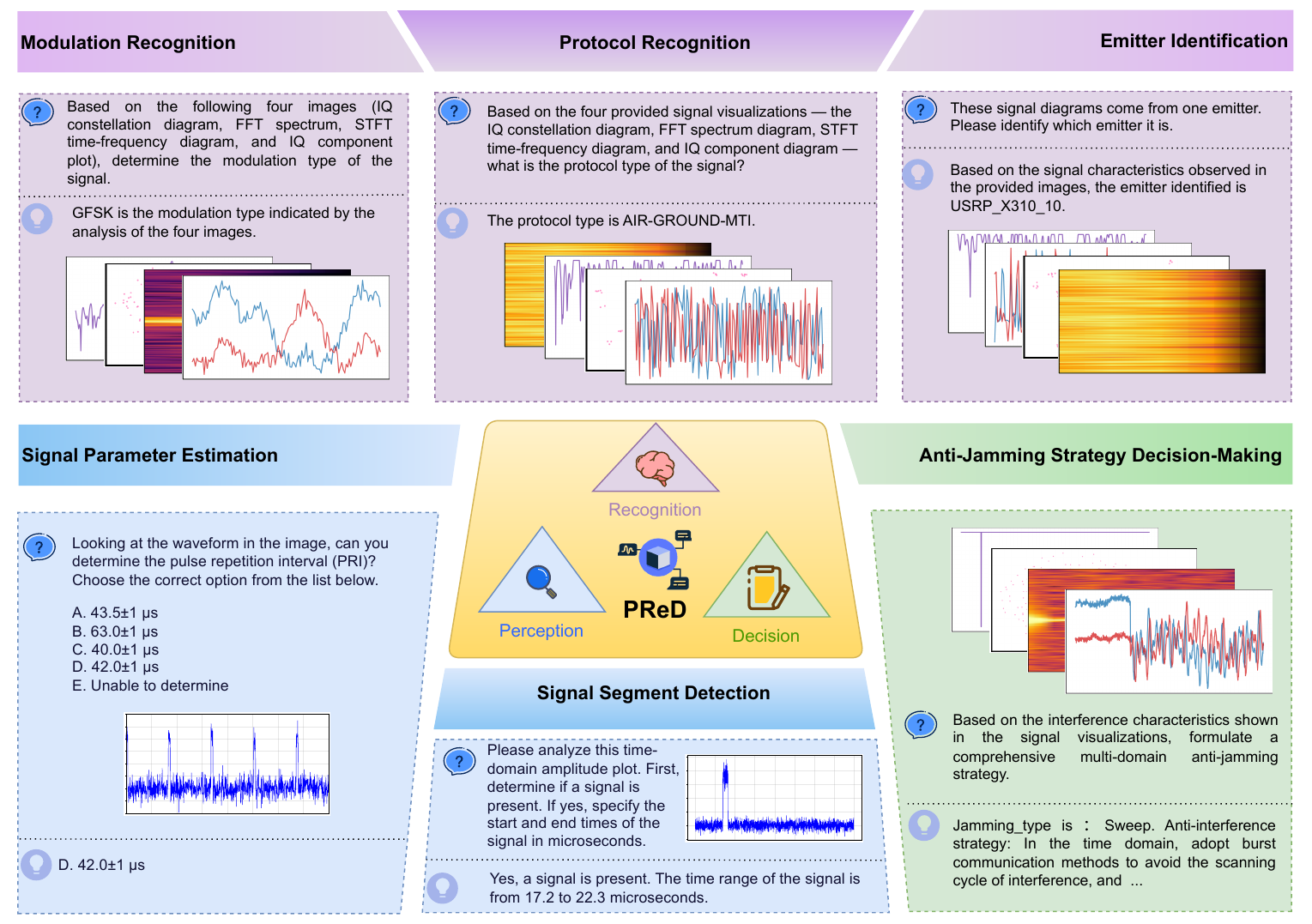}
     \caption{Overview of PReD and its capabilities, instantiated across six core electromagnetic (EM) tasks that span the three tiers of perception, recognition, and decision-making. The figure demonstrates how PReD processes multi-view signal visualizations to address a diverse range of EM instructions, from quantitative parameter estimation to open-ended anti-jamming strategy generation.}
     \label{fig:pred_overview}
 \end{figure}
 }]    
    
 \begin{abstract}

    \itshape 
    Multimodal Large Language Models have demonstrated powerful cross-modal understanding and reasoning capabilities in general domains. However, in the electromagnetic (EM) domain, they still face challenges such as data scarcity and insufficient integration of domain knowledge. This paper proposes PReD, the first foundation model for the EM domain that covers the intelligent closed-loop of "perception, recognition, decision-making." We constructed a high-quality multitask EM dataset, PReD-1.3M, and an evaluation benchmark, PReD-Bench. The dataset encompasses multi-perspective representations such as raw time-domain waveform, frequency-domain spectrograms, and constellation diagrams, covering typical features of communication and radar signals. It supports a range of core tasks, including signal detection, modulation recognition, parameter estimation, protocol recognition, radio frequency fingerprint recognition, and anti-jamming decision-making. PReD adopts a multi-stage training strategy that unifies multiple tasks for EM signals. It achieves closed-loop optimization from end-to-end signal understanding to language-driven reasoning and decision-making, significantly enhancing EM domain expertise while maintaining general multimodal capabilities. Experimental results show that PReD achieves state-of-the-art performance on PReD-Bench constructed from both open-source and self-collected signal datasets. These results collectively validate the feasibility and potential of vision-aligned foundation models in advancing the understanding and reasoning of EM signals.
\end{abstract}


 \section{Introduction}
\label{sec:intro}

Recent multimodal large language models (MLLMs), enabled by instruction tuning and cross-modal alignment, have achieved notable breakthroughs on visual question answering, retrieval, and reasoning tasks (e.g., LLaVA \cite{Liu2023VisualInstructionTuning}, BLIP-2 \cite{Li2023BLIP2}, GPT-4V \cite{OpenAI2023GPT4VSystemCard}, ImageBind \cite{Girdhar2023ImageBind}), driving a unified intelligence paradigm that uses natural language as the primary interface. However, these general-purpose large language models (LLMs) principally rely on world knowledge and language-reasoning capabilities derived from image, video, and audio modalities; they lack intrinsic representations of electromagnetic (EM) signal priors such as raw In-phase and quadrature (I/Q) waveforms, spectrogram structures, constellation geometry, micro-Doppler signatures, and cyclostationarity. This deficit limits their applicability in high-demand domains such as spectrum management, wireless security, and radar semantic understanding, where expert-level, auditable reasoning about EM signals is required. Meanwhile, EM-/wireless-oriented LLM efforts have emerged (e.g., WirelessLLM \cite{Shao2024WirelessLLM}, RadioLLM \cite{Chen2025RadioLLM}, RFF-LLM \cite{Zheng2025UAVIDLLM}), but many of these concentrate on network planning \cite{DBLP:journals/ieeenl/QuanNZYXWLS25, DBLP:conf/sigcomm/WangZLL0ZAWG0CL25} or knowledge-level QA \cite{DBLP:conf/comsnets/NatarajanDB25} and treat signal features as external prompts rather than forming a unified foundation for raw EM data. Together, these observations indicate that an integrated ``model--data--evaluation'' framework tailored to the shapes and task ecology of EM data is essential.

A primary challenge in the EM domain is constructing datasets suitable as inputs to large EM models. Public EM signal repositories are scarce or insufficient: many are generated from simulations or cover single communication scenarios, with limited diversity in modulation formats, protocol stacks, and interference types \cite{qin2025scaling}. Representative datasets such as RadioML2016 \cite{OShea2016ConvRadio}, RadioML2018 \cite{OShea2018OverTheAir}, and HisarMod2019.1 \cite{Tekbiyik2020AMC} are widely used in research but are not directly fit for training and evaluating EM large models. Second, unified multi-task capability is lacking: tasks including signal detection, modulation recognition, parameter estimation, protocol classification, radio frequency (RF) fingerprinting, and anti-jamming strategy selection are typically modeled separately, and there is no widely adopted base that unifies these tasks in a shared multimodal EM feature space aligned with instruction-driven interfaces. Third, a clear expert gap exists when transferring general-purpose MLLMs to the EM domain: a general-purpose LLMs often fails to internalize EM priors (e.g., cyclostationarity, carrier/symbol timing, inter-band coupling), yielding unstable perception and reasoning under low SNR, strong interference, or complex waveforms, and limiting interpretability.

Existing EM large-model research typically focuses on single-task or knowledge-centric intelligence. For example, WirelessLLM emphasizes knowledge alignment, rule understanding, and network optimization but rarely operates directly on raw I/Q waveforms or spectrograms \cite{Shao2024WirelessLLM}. RadioLLM integrates LLMs in cognitive radio via prompt and token reprogramming \cite{Chen2025RadioLLM}, yet it remains largely a ``language–feature prompt'' approach and doesn’t deliver an end-to-end perceptual-to-decisional capability from signals to semantic actions. In classical RF machine learning, deep models have advanced modulation classification, time–frequency feature modeling, synchronization/registration, and protocol-sequence recognition, but these successes are typically confined to task-specific, small-scale models with limited data and lack instruction alignment and standardized evaluation that would support closed-loop perception, recognition, and decision-making generalization and verifiability \cite{OShea2016UnsupervisedRep,OShea2016RTN,OShea2016RNNRadio}.

We present a unified foundation model for EM perception, recognition, and decision-making, termed \textbf{PReD}. An overview of our entire framework is provided in Fig~\ref{fig:pred_overview}. At the data layer, we construct multi-dimensional, feature-visualized signal corpora that raw I/Q waveform, frequency-domain spectrum, short-time Fourier transform (STFT) spectrograms, and constellation diagrams. Eight human annotators provide reasoning-style descriptions and consistency labels under a unified question–answer protocol, forming multi-granularity supervision for multiple task families, including signal segment detection (SSD), signal parameter estimation (SPE), modulation recognition (MR), protocol recognition (PR), emitter identification (EI), and anti-jamming strategy decision-making (AJSD). At the model layer, we design a visual input paradigm that jointly represents EM signals from complementary views across time, frequency, and constellation domains, and adopt a multi-stage training pipeline to support multi-task learning at scale. At the learning paradigm level, we introduce unified multi-task instruction alignment that uses language as a cross-task interface, enabling an end-to-end loop from perception to recognition and to decision-making. This design retains the cross-modal expressiveness of general multimodal large language models while compensating for missing EM priors.

The main contributions are as follows:
\begin{itemize}
    \item To the best of our knowledge, this work proposes and implements the first integrated foundation model for the EM domain that unifies perception, recognition, and decision-making.
    \item To support efficient training and comprehensive evaluation, we build a large-scale, cross-view EM dataset that covers time, frequency, time-frequency, and constellation perspectives with reasoning-style supervision.
    \item Extensive experiments indicate competitive performance across diverse EM tasks; datasets, models, and related code will be released to facilitate further research.
\end{itemize}

 \section{Related Work}

Recent years have seen sustained advances in Large Language Models (LLMs) and multimodal variants (MLLMs) \cite{ji2025robobrain}. Via instruction tuning, cross-modal alignment, and tool augmentation, these systems advance general-purpose intelligence in visual question answering, retrieval, reasoning, and open-world understanding \cite{tan2025reason}. Representative image–text models include BLIP-2 \cite{Li2023BLIP2}, LLaVA \cite{Liu2023VisualInstructionTuning}, MiniGPT-4 \cite{Zhu2023MiniGPT4}, and Qwen-VL \cite{Bai2023QwenVL}; unified cross-sensor representations are explored by ImageBind \cite{Girdhar2023ImageBind}; long-context video–text and tool integration are addressed by Flamingo \cite{Alayrac2022Flamingo}, Kosmos-2 \cite{Peng2023Kosmos2}, and GPT-4Vis \cite{Wu2024GPT4Vis}. Together, these works construct transferable interfaces among natural images, audio and speech, web corpora, and language patterns, enabling a language-centered multitask paradigm \cite{wang2025towards}.

\subsection{General-Purpose Multimodal Large Models}

A prevailing design treats an LLM as the semantic hub coupled with vision and audio encoders. Cross-modal alignment is realized through projection heads or query routers, and instruction tuning with preference optimization harmonizes multitask behavior. This ``backbone–alignment–instruction'' pipeline has become a standard for reasoning over images, video, and speech \cite{ji2025visualtrans}.
\textit{Transfer to EM is nontrivial}: (i) missing modality priors—cyclostationarity, carrier/symbol timing offsets, phase–envelope and constellation geometry, inter-band spectral coupling, and micro-Doppler—are not captured by features from natural image or audio corpora \cite{Khalek2024CognitiveRadioSurvey,He2019ModelDrivenDL}; (ii) interface mismatch—beyond categorical recognition, EM requires regression and closed-loop control (e.g., SNR and timing-offset estimation, channel/jammer recovery, synchronization, anti-jamming search), whereas generic instruction spaces emphasize QA-style generation \cite{ji2025mathsticks}; (iii) supervision/metric mismatch—datasets seldom provide executable physical ground truths or counterfactual controls, encouraging semantic plausibility over testable EM correctness.

\subsection{Large-Model Directions in the EM Domain}

\paragraph{Knowledge alignment and cognitive-radio-oriented work.}
WirelessLLM targets network planning, rule understanding, and knowledge-based QA for cognitive wireless decisions but seldom operates directly on raw I/Q waveforms or spectrograms \cite{Shao2024WirelessLLM}. RadioLLM introduces prompt and token reprogramming to embed LLMs in cognitive-radio workflows and improves programmability, yet it remains a language–external-feature paradigm without unified modeling of raw EM data \cite{Chen2025RadioLLM}. RFF-LLM enhances device-level interpretability via language interfaces but is limited by single-task scope and controlled data \cite{Zheng2025UAVIDLLM}.
\textit{Synthesis of limitations}: existing approaches rarely cover, within one framework, signal detection, modulation/protocol recognition, emitter identification, parameter estimation, and anti-jamming decision-making under a unified instruction space; moreover, EM-derived features are often used as auxiliary prompts rather than a joint representation enforcing cross-view consistency among I/Q, spectrum, spectrograms, and constellations, which weakens robustness in low-SNR regimes. These gaps motivate EM-tailored foundation models that integrate data, modeling, and evaluation to close the loop across perception, recognition, and decision-making for auditable EM intelligence \cite{ji2025enhancing}.

\begin{figure*}[t]
    \centering
    \includegraphics[width=0.95\textwidth]{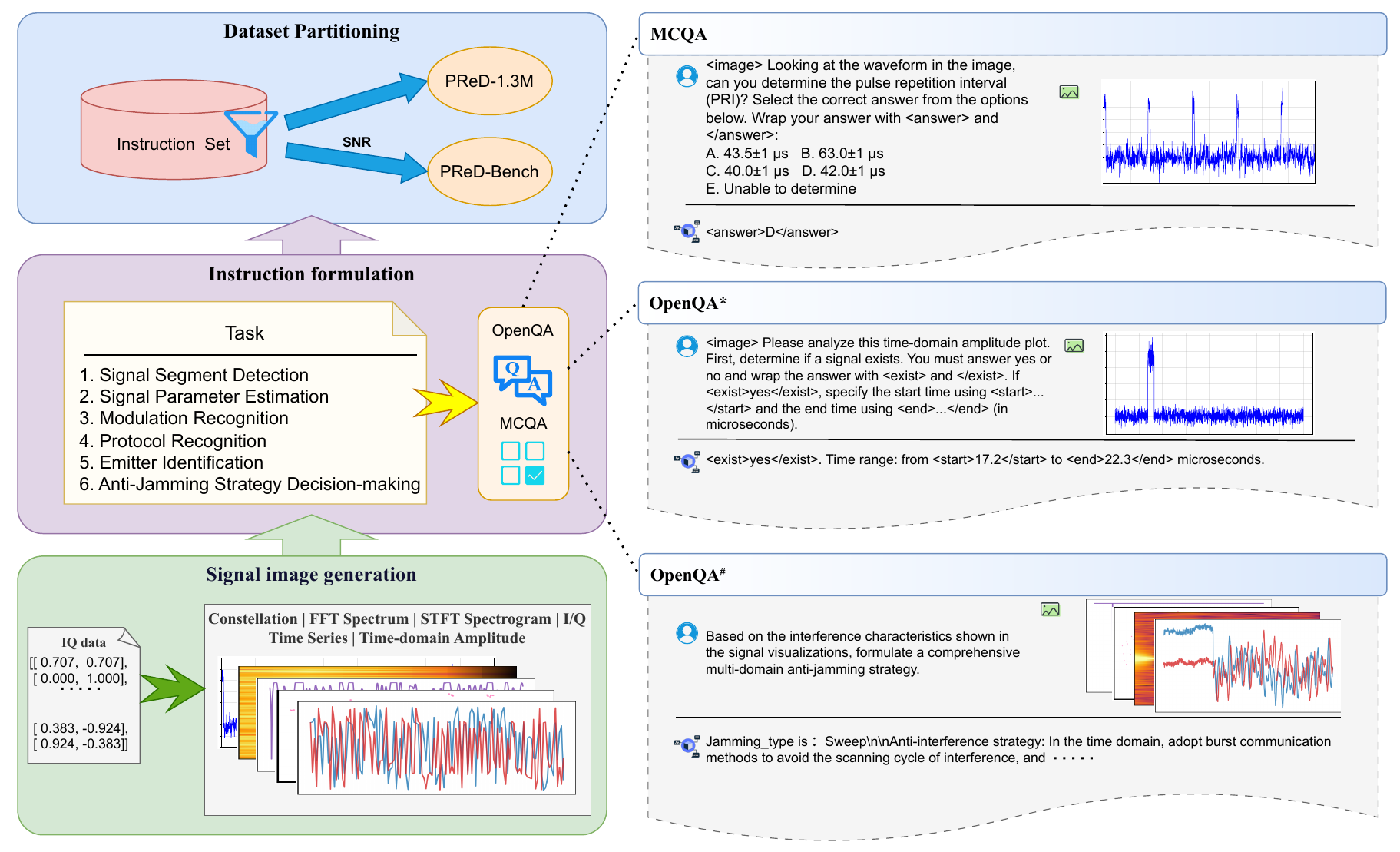}
    \caption{Overall pipeline for the construction of the PReD-1.3M training set and the PReD-Bench. We first generate five types of signal visualizations from the raw IQ signals. We then create corresponding OpenQA and MCQA pairs for each task type to form a complete instruction set. Finally, a portion of this set is sampled to obtain the PReD-Bench, which is kept strictly separate from the training data.}
    \label{fig:EM_Dataset_Construction_Pipeline}
\end{figure*}

\section{The PReD Dataset and Benchmark}
In this section, we detail the construction of our large-scale EM instruction corpus, PReD-1.3M, and our hierarchical evaluation benchmark, PReD-Bench. Figure~\ref{fig:EM_Dataset_Construction_Pipeline} illustrates the overall pipeline, which begins with signal visualization, proceeds to instruction formulation, and concludes with the partitioning of the instruction set into training and evaluation splits. We begin by describing the shared methodology for data formulation.
\subsection{Signal Representation and Task Formulation}

The EM instruction formulation defines six core tasks: SSD, SPE, MR, PR, EI, and AJSD. Each instruction is paired with visual signal representations such as waveform, FFT spectrum, STFT spectrogram, or constellation images. Multi-format question–answer templates (OpenQA and MCQA) are used to synthesize benchmark-ready supervision for MLLMs training.

\begin{figure*}[t]
    \centering
    \begin{subfigure}[t]{0.48\textwidth}
        \centering
        \includegraphics[width=\linewidth]{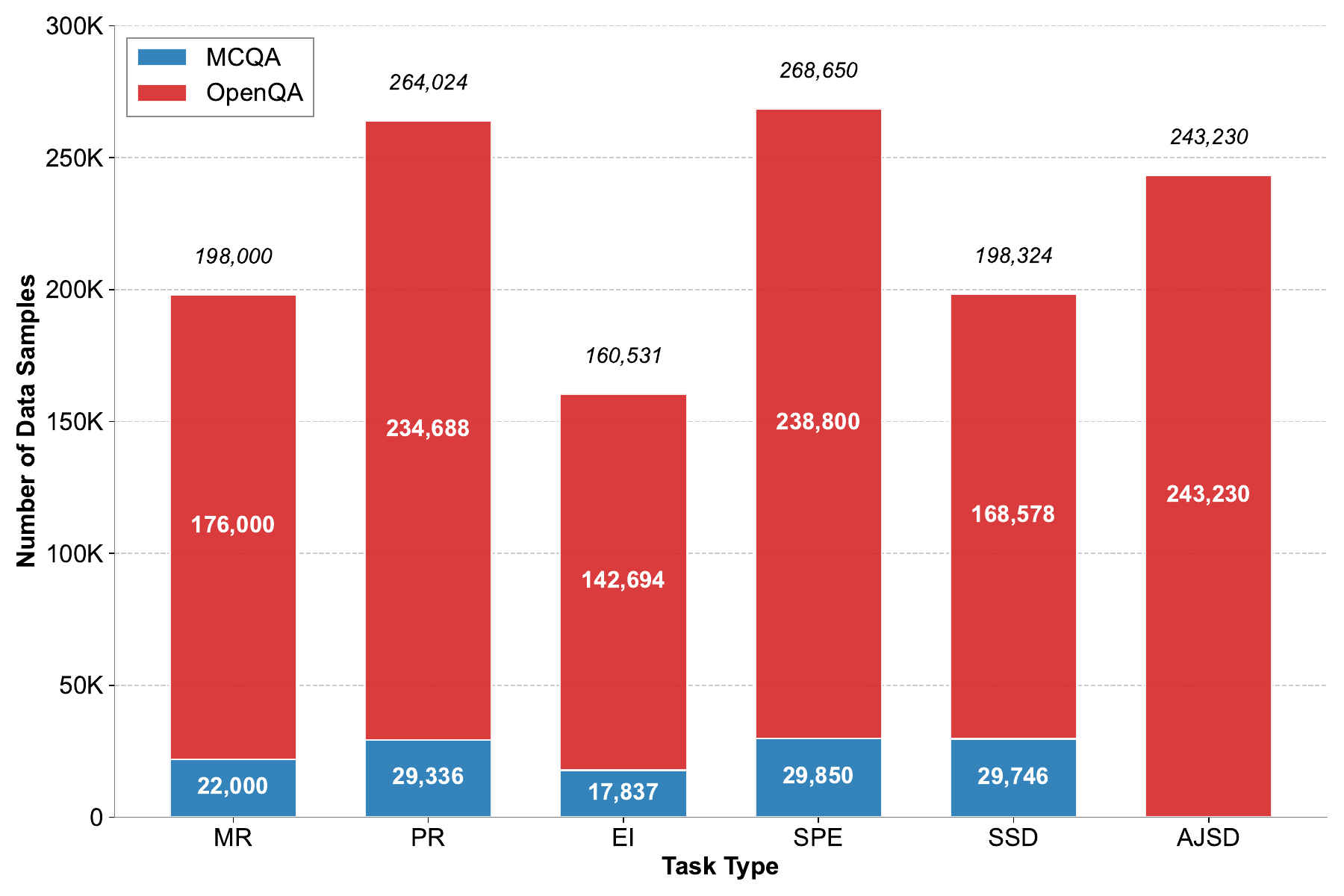}
        \caption{Task-wise sample counts split by OpenQA (red) and MCQA (blue).}
        \label{fig:em_task_counts}
    \end{subfigure}\hfill
    \begin{subfigure}[t]{0.48\textwidth}
        \centering
        \includegraphics[width=\linewidth]{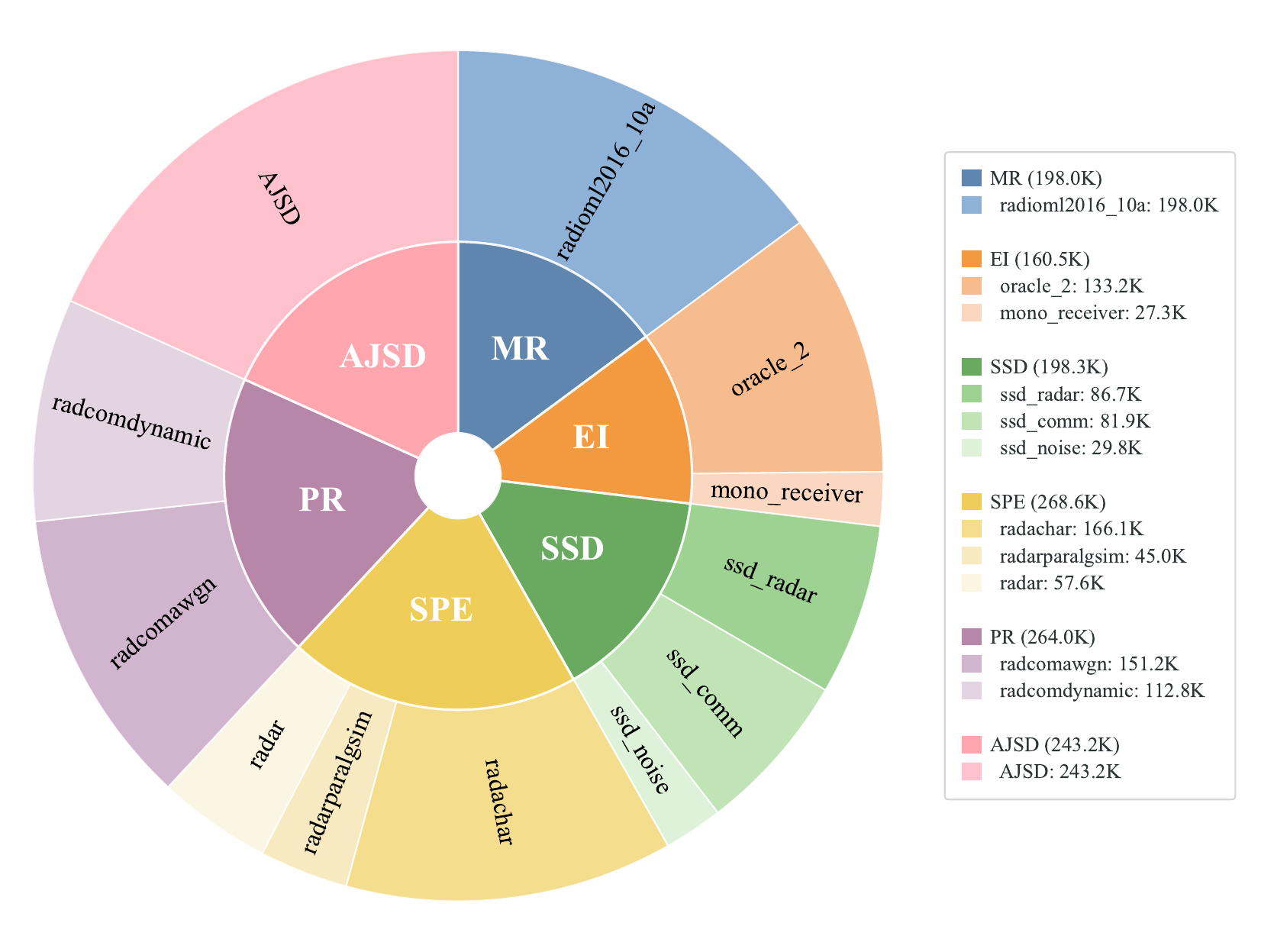}
        \caption{Source composition per task (outer ring) organized by six top-level tasks (inner ring).}
        \label{fig:em_task_sources}
    \end{subfigure}
    \caption{Scale and composition of the EM training dataset across six tasks and two QA formats.}
    \label{fig:em_bench_twofigs}
\end{figure*}

An EM signal is fundamentally a complex-valued time series \(x[n]\in\mathbb{C}^N\). We map each raw sample to four complementary visual views:
\begin{itemize}
    \item \textbf{I/Q Constellation} — sampled scatter points \( \{x[k]\} \) that reveal symbol constellations and modulation geometry;
    \item \textbf{FFT Magnitude Spectrum} — \(S_{\mathrm{FFT}}(f)=|\mathrm{FFT}(x[n])|\), capturing bandwidth and frequency-domain structure;
    \item \textbf{STFT Spectrogram} — \(S_{\mathrm{STFT}}(t,f)\), trading off time–frequency resolution to reveal pulsed or sweep patterns;
    \item \textbf{Raw I/Q Waveform} — real and imaginary component trajectories \(x_I[n]\) and \(x_Q[n]\), together with the amplitude envelope for parameter readout.
\end{itemize}

To support instruction-style training \cite{DBLP:conf/iclr/WeiBZGYLDDL22}, we design two question formats and corresponding output tags. Each signal sample in MCQA format includes a built-in \texttt{"Unable to answer"} option to mitigate model overconfidence on ambiguous inputs.
\begin{itemize}
    \item \textbf{Multiple Choice Question-Answer Pairs:} the \texttt{user} presents four images and five options (including \texttt{Unable to answer}); the \texttt{assistant} returns \texttt{<answer>A</answer>} \cite{DBLP:conf/naacl/HuangWZGZ022}.
    \item \textbf{Open-ended Question-Answer Pairs:} the \texttt{user} provides four images and instructs ``to only output the result''; the \texttt{assistant} responds with a structured tag such as \texttt{<mode>…</mode>}, facilitating automated scoring and downstream integration \cite{DBLP:conf/emnlp/MihaylovCKS18}.
\end{itemize}

\subsection{The PReD-1.3M Corpus}
To facilitate the domain adaptation of our model, we construct \textbf{PReD-1.3M}, a large-scale EM instruction corpus. As summarized by the task-wise histogram in Fig.~\ref{fig:em_task_counts}, the corpus totals \(\mathbf{1.33\times10^6}\) question–answer pairs. OpenQA supervision constitutes the majority (approximately \(\mathbf{90.3\%}\)), with the remaining \(\sim\!9.7\%\) provided in the MCQA format. The sunburst chart in Fig.~\ref{fig:em_task_sources} details the source composition for each task, demonstrating a rich mix of public and controllable synthetic corpora. The corpus is broken down by task as follows:

\paragraph{(A) Signal Segment Detection.}
We aggregate three sources—\texttt{radar}, \texttt{communication}, and \texttt{noise}—to form a supervised corpus of short complex-baseband (IQ) segments acquired at \(20\,\mathrm{MS/s}\) with SNR spanning \([-10, 20]\,\mathrm{dB}\). The noise-only subset is retained as hard negatives. The dataset comprises \(\approx 1.98\times 10^5\) labeled segments.

\paragraph{(B) Signal Parameter Estimation.}
The parameter estimation task is cast into an instruction format that requests quantities such as pulse width, period, count, and time delay. The input is a single time-domain amplitude trace, and the correct MCQA choice is provided as \(\text{value}\pm 1~\mu\text{s}\). Distractors are generated using multiplicative, offset, and random-sampling strategies, and options that are too close to the ground truth are removed to increase separability. The parameter-estimation set contains 269K samples (OpenQA: 239K; MCQA: 30K), spanning multiple \(f_s\) settings (e.g., 10/12 MHz) and approximately covering \([-20, +20]\) dB in SNR.

\paragraph{(C) Modulation Recognition.}
The modulation recognition task is built on RadioML2016.10a. We fix \(f_s=1\) MHz and sample SNR uniformly over \([-20, +18]\) dB. Modulation classes include AM-DSB/SSB, BPSK/QPSK/8PSK, QAM16/64, GFSK, CPFSK, PAM4, WBFM, etc. Each sample is rendered into four images, and two instruction types (OpenQA and MCQA) are generated. The modulation dataset totals 198K samples (OpenQA: 176K; MCQA: 22K).

\paragraph{(D) Protocol Recognition.}
The protocol dataset covers avionics ranging, surveying, beaconing, short-range protocols, Bluetooth, IEEE 802.11 / 802.15.4, and other protocols. The typical sampling rate is \(f_s=10\) MHz. Examples are generated via instruction-style QA. The constructed protocol dataset contains 263K samples (OpenQA: 234K; MCQA: 29K). SNRs are sampled within \([-20, 18]\) dB.

\paragraph{(E) Emitter Identification.}
For emitter identification, we collect real and simulated receiver-chain variations to create device fingerprints (Oracle\_2 / Mono-Receiver basis). Device names are prefixed in the templates (e.g., \texttt{USRP B210/X310 N}), and \texttt{Unable\_to\_answer} is retained to filter ambiguous boundary cases. The EI dataset comprises 161K samples (OpenQA: 143K; MCQA: 18K), covering multiple sampling rates and a long-tailed device distribution.

\paragraph{(F) Anti-Jamming Strategy Decision-Making.}
We include pure-noise examples, single- and multiple-interferer regimes, linear frequency modulation, phase-coded waveforms, and diverse radar modes. We provide causal OpenQA pairs for AJSD task to strengthen the model's strategy decision-making capability.


\begin{table}[t]
    \centering
    \begin{tabular}{cccc}
        \hline
        Dimension & Task & OpenQA & MCQA \\
        \hline
        \multirow{2}{*}{EM Perception}
        & SPE  & 2250   & 750  \\
        & SSD  & 1700   & 300  \\
        \cline{2-4}
        \multirow{3}{*}{EM Recognition}
        & MR   & --     & 500  \\
        & PR   & --     & 500  \\
        & EI   & --     & 458  \\
        \cline{2-4}
        EM Decision-making
        & AJSD & 2000   & --   \\
        \hline
    \end{tabular}
    \caption{PReD-Bench for perception, recognition, and decision-making. 
    Abbreviations: SSD = Signal Segment Detection; SPE = Signal Parameter Estimation;
    MR = Modulation Recognition; PR = Protocol Recognition; EI = Emitter Identification;
    AJSD = Anti-Jamming Strategy Decision-making. OpenQA and MCQA denote open-ended and multiple-choice question–answer pairs, respectively.}
    \label{tab:dataset_composition}
\end{table}

\subsection{The PReD-Bench Benchmark}

To systematically evaluate general-purpose models in the EM domain, we introduce a hierarchical EM domain benchmark. The benchmark is constructed via SNR-stratified sampling and is kept strictly isolated from the PReD-1.3M training set to prevent any data leakage. It organizes the perception–recognition–decision-making pipeline into six representative tasks, with unified I/O conventions and scoring for both OpenQA and MCQA. Table~\ref{tab:dataset_composition} specifies the exact per-task item counts by question type. Data are curated from multi-origin public sets and controllable simulation, followed by normalization and quality auditing. Prompts and rubric-based scoring are standardized to enable cross-task comparability and reproducible evaluation, and anti-jamming strategy decision-making is explicitly included to stress closed-loop reasoning under adversarial EM conditions \cite{ji2026prm}.

 \section{PReD Model}

\begin{figure*}[t]
    \centering
    \includegraphics[width=\textwidth]{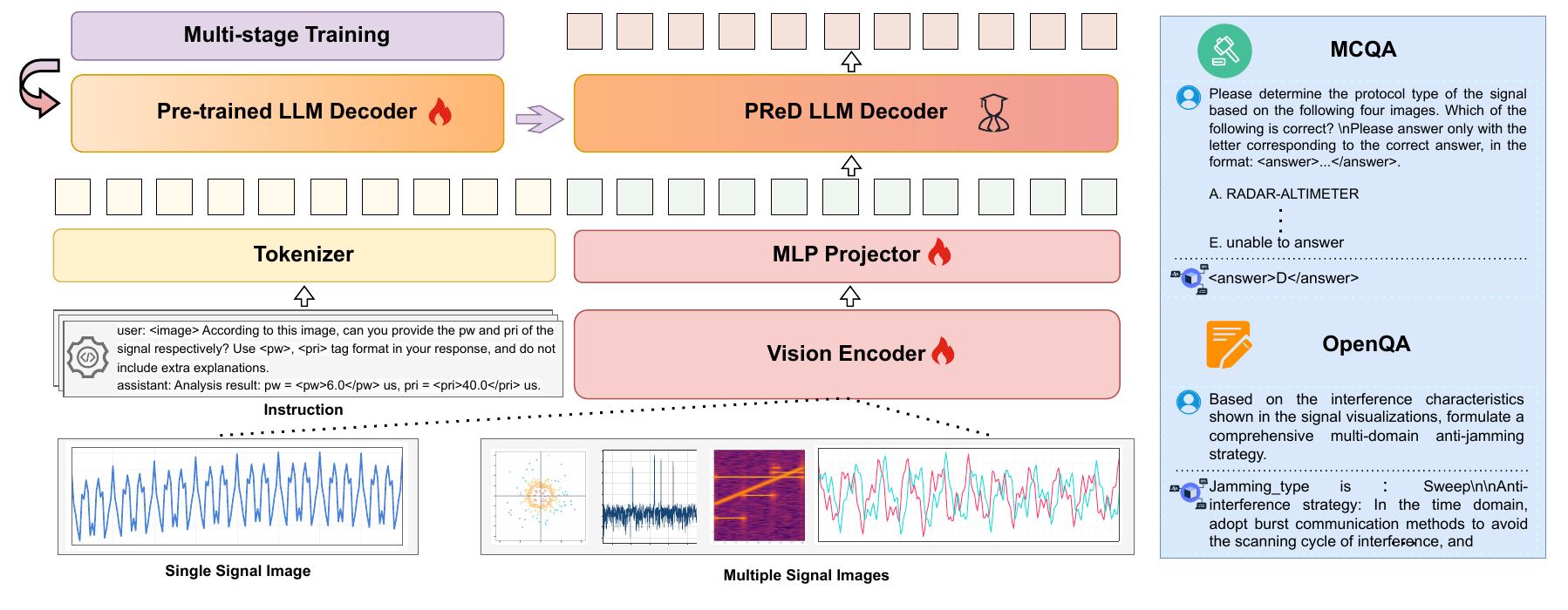}
    \caption{
        Overall pipeline of PReD. (Bottom to top) Multi-view EM renderings are encoded and projected into the token space, where they are combined with tokenized instructions. A pre-trained LLM decoder is first aligned with visual features and then fine-tuned through a multi-stage curriculum to become the specialized \emph{PReD LLM Decoder}. The model supports two output modes: open-ended generation (OpenQA) and structured multiple-choice question answering (MCQA).}
    \label{fig:pred-framework}
\end{figure*}

\subsection{Model Architecture}
\label{sec:model_architecture}

PReD adopts a standard yet effective multimodal architecture inspired by LLaVA~\cite{llava2023}, comprising three key components: a vision encoder, a projector, and a large language model decoder.

\noindent\textbf{Core Components.} We employ \textbf{SigLIP} as the vision encoder ($f_v$) to extract features from each visual view ($X_k$). A two-layer MLP projector ($g$) then maps these visual tokens into the embedding space of the language model. For the decoder, we utilize the pretrained \textbf{Qwen3-8B} \cite{qwen3technicalreport} model ($\phi$).

\noindent\textbf{Multi-View and Text Integration.} To handle multiple views, we follow a standard packing strategy where visual tokens from each view are concatenated, separated by a learnable boundary token, and augmented with a view-index embedding \cite{DBLP:conf/icml/RadfordKHRGASAM21}. This packed visual prefix, denoted as $\Pi(\mathcal{X})$, is then prepended to the tokenized instruction prompt $P$. The entire sequence is processed by the single autoregressive Qwen3-8B decoder, which models the probability of the response sequence $Y$ as shown in Eq.~\eqref{eq:autoregressive}. Our training methodology, which involves progressively unfreezing and fine-tuning these components, is detailed in Section~\ref{sec:multi_stage_training}. 

 \begin{equation}
     p_\phi(Y\mid C)=\prod_{t=1}^{L}p_\phi\!\left(y_t~\middle|~C,\,y_{<t}\right),
     \label{eq:autoregressive}
 \end{equation}
 
\noindent\textbf{Training Objective.}
We define the language modeling loss
\begin{equation}
    \mathcal{L}_{\mathrm{LM}}(\phi)
    = -\sum_{t=1}^{L}\log p_{\phi}\!\left(y_t~\middle|~C,\,y_{<t}\right).
    \label{eq:lm_loss}
\end{equation}

\begin{table*}[t]
    \centering
    \begin{minipage}{0.9\textwidth}
        \centering
        \includegraphics[width=1.05\textwidth]{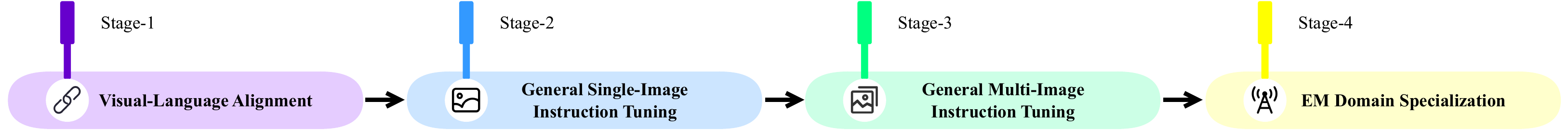}\\[4pt]

        \setlength{\tabcolsep}{6pt}
        \renewcommand{\arraystretch}{1.25}
        \resizebox{\textwidth}{!}{%
            \begin{tabular}{l|l|c|c|c|c}
                \hline
                Category & Hyperparameter        & Stage-1          & Stage-2                   & Stage-3                   & Stage-4                   \\
                \hline
                \multirow{2}{*}{Vision}
                & Resolution            & $384$             & Max $384\times(2\times2)$ & Max $384\times(6\times6)$ & Max $384\times(6\times6)$ \\
                & Tokens                & $729$             & Max $729\times5$          & Max $729\times10$         & Max $729\times10$         \\
                \hline
                \multirow{2}{*}{Data}
                & Dataset               & LCS               & General Single-Image      & General Multi-Image*      & \textbf{PReD-1.3M}*       \\
                & Samples               & $558\mathrm{K}$   & $858\mathrm{K}$           & $832\mathrm{K}$           & $1.8\mathrm{M}$           \\
                \hline
                \multirow{2}{*}{Model}
                & Trainable Components  & Projector         & Full Model                & Full Model                & Full Model                \\
                & Tunable Parameters    & $21.5\mathrm{M}$  & $8.6\mathrm{B}$           & $8.6\mathrm{B}$           & $8.6\mathrm{B}$           \\
                \hline
                \multirow{4}{*}{Training}
                & Per-Device Batch Size & $8$               & $4$                       & $2$                       & $2$                       \\
                & LR: $\psi_{ViT}$      & N/A               & $2\times10^{-6}$          & $2\times10^{-6}$          & $2\times10^{-6}$          \\
                & LR: $\{\theta_{\mathrm{Proj}},\ \phi_{\mathrm{LLM}},\ \phi_{\mathrm{LoRA}}\}$ 
                & $1\times10^{-3}$      & $1\times10^{-5}$  & $1\times10^{-5}$          & $1\times10^{-5}$          \\
                & Epochs                & $1$               & $1$                       & $1$                       & $1$                       \\
                \hline
            \end{tabular}
        }
    \end{minipage}

    \caption{Training schedule for PReD across four stages on 16 NVIDIA A100 GPUs (80GB each). A star * indicates datasets mixed with general visual corpora from earlier stages.}
    \label{tab:hyperparams_stages}
\end{table*}

\subsection{Multi-Stage Training}
\label{sec:multi_stage_training} 
To couple general multimodal competence with EM expertise, we train PReD with a four-stage curriculum.

\noindent\textbf{Stage-1: Projector Warm-Up.}
We freeze $\phi_0$, $f_v$ and train $g$ on single-view inputs to align vision tokens at a $384$ resolution and $729$ vision-token length.

\noindent\textbf{Stage-2: General Single-View.}
We unfreeze the full model and train with general single-image corpora, using up to $5\times 729$ tokens to consolidate instruction following.

\noindent\textbf{Stage-3: General Multi-View.}
We introduce multi-view inputs (up to 
$10\times 729$ tokens) to develop cross-view reasoning. To preserve single-view competence, we adopt the LLaVA-Next interleaving strategy, mixing the new data with 40\% of the Stage-2 corpus.

\noindent\textbf{Stage-4: EM Domain Adaptation.}
Finally, for domain adaptation, the model is fine-tuned on our PReD-1.3M dataset, covering all six EM tasks. To retain generalist skills, we continue this interleaving approach by mixing the specialized data with 25\% of the Stage-2 single-view and 50\% of the Stage-3 multi-view corpora. Training schedules follow Table~\ref{tab:hyperparams_stages}.

 \section{Experiment and Results}

\subsection{Performance Evaluation of EM Tasks}


Table~\ref{tab:signal-benchmark} demonstrates that across the perception–recognition–decision-making pipeline, PReD consistently outperforms all general-purpose baselines and remains strong in the decision-making stage that requires untagged, generative matching (\texttt{OpenQA}). The results reveal a clear capability boundary: while general-purpose models show rudimentary skill on simple perception tasks—for instance, GPT-5 achieves 61.0 accuracy on the SSD (MCQA) task—they completely fail at complex recognition and decision-making. This is most evident in the anti-jamming (AJSD) task, where all generalist models score near zero on BLEU4, proving incapable of generating valid strategies. In contrast, PReD excels (BLEU4=0.253). This confirms that generalist models are limited to shallow pattern matching due to a lack of domain priors, whereas PReD successfully forges the full perception-to-decision pipeline via its specialized architecture.


\begin{table*}[htbp]
    \centering
    \setlength{\tabcolsep}{3.5pt}
    \renewcommand{\arraystretch}{1.15}
    \resizebox{\textwidth}{!}{%
        \begin{tabular}{@{}l*{11}{c}@{}}
            \toprule
            \multirow{3}{*}{Model}
            & \multicolumn{4}{c}{\textbf{EM Perception}}
            & \multicolumn{3}{c}{\textbf{EM Recognition}}
            & \multicolumn{4}{c}{\textbf{EM Decision-Making}} \\
            \cmidrule(lr){2-5}\cmidrule(lr){6-8}\cmidrule(lr){9-12}
            & \multicolumn{2}{c}{SSD} & \multicolumn{2}{c}{SPE}
            & \multicolumn{1}{c}{MR} & \multicolumn{1}{c}{PR} & \multicolumn{1}{c}{EI}
            & \multicolumn{4}{c}{\makecell[c]{AJSD (\texttt{OpenQA$^\#$})}} \\
            \cmidrule(lr){2-3}\cmidrule(lr){4-5}\cmidrule(lr){6-8}\cmidrule(lr){9-12}
            & MCQA & \texttt{OpenQA$^*$}
            & MCQA & \texttt{OpenQA$^*$}
            & MCQA & MCQA & MCQA
            & BLEU4 & ROUGE & METEOR & CIDEr \\
            \midrule
            \makecell[l]{qwen2.5-vl-7b$\ssymbol{2}$ \cite{qwen2.5-VL}}       & 28.7 & 0.2  & 20.5 & 0.1  & 24.2 & 14.0 & --   & 0.000 & 0.079 & 0.115 & 0.128 \\
            \makecell[l]{qwen3-vl-8b$\ssymbol{2}$ \cite{qwen3technicalreport}}         & 21.3 & 1.3  & 18.8 & 2.0  & 11.6 & 2.8  & --   & 0.000 & 0.094 & 0.121 & 0.228 \\
            \makecell[l]{qwen3-vl-235b-a22b$\ssymbol{2}$ \cite{qwen3technicalreport}}  & 27.7 & 0.5  & 30.4 & 2.0  & 20.2 & 8.6  & --   & 0.000 & 0.090 & 0.070 & 0.151 \\
            \makecell[l]{seed-1.6 vision \cite{guo2025seed15vltechnicalreport}}                & 41.3 & 1.5  & 40.4 & 3.1  & 23.4 & 21.4 & --   & 0.000 & 0.095 & 0.039 & 0.169 \\
            \makecell[l]{claude sonnet-4 \cite{anthropic2024claude}}                  & 28.7 & 1.4  & 48.5 & 2.8  & 25.2 & 21.6 & --   & 0.000 & 0.103 & 0.078 & 0.154 \\
             \makecell[l]{GPT-5 \cite{openai2025gpt5}}
                                                    & 61.0 & 7.2  & 42.9 & 13.3 & 12.0 & 19.2 & --   & 0.000 & 0.017 & 0.028 & 0.055 \\
            \makecell[l]{gemini-2.5 pro \cite{comanici2025gemini25pushingfrontier}}                  & 25.3 & 6.2  & 56.3 & 12.0 & 29.0 & 24.0 & --   & 0.000 & 0.081 & 0.036 & 0.129 \\
            \textbf{PReD (Ours)}                          & \textbf{99.3} & \textbf{80.6} & \textbf{97.1} & \textbf{81.9} & \textbf{72.0} & \textbf{72.0} & \textbf{69.4} & \textbf{0.253} & \textbf{0.559} & \textbf{0.515} & \textbf{0.617} \\
            \bottomrule
        \end{tabular}

    }

    \caption{Benchmark results for EM perception, recognition, and decision-making tasks. A superscript \(\ssymbol{2}\) marks an Instruct (instruction-tuned) variant.
    \texttt{OpenQA$^*$} indicates answers with a specific tag whose results are directly matched;
    \texttt{OpenQA$^\#$} indicates answers without a specific tag, where matching is performed using BLEU4, ROUGE, METEOR, and CIDEr.
    Since EI task requires device-specific RF fingerprint learning, general MLLMs arenot comparable and their results are omitted.}
    \label{tab:signal-benchmark}
\end{table*}

\subsection{Evaluation on General Multimodal Benchmarks}
An important question is whether PReD’s domain-oriented optimization affects its general multimodal understanding capabilities. To investigate this, we conducted comprehensive evaluations on several widely used benchmarks (see Table~\ref{tab:gen_mm_bench}). Our main comparisons include representative open-source models of similar scale, with LLaVA-Next-Interleave-7B serving as a key methodological reference.

The results show that PReD achieves an average score of 65.4, slightly surpassing its reference model LLaVA-Next (64.0) and on par with other strong open-source baselines such as Qwen-VL-Max (65.2). This indicates that PReD successfully maintains competitive general multimodal understanding while benefiting from its domain-oriented enhancements.

For a broader perspective, we also report results from the high-performance proprietary reference GPT-4V (1106-preview, Nov. 2023). PReD demonstrates comparable overall capability (65.4 vs. 66.9) and notably achieves stronger robustness on HallusionBench (54.5 vs. 46.5). This suggests that our proposed training paradigm may contribute to improved reliability under hallucination-prone scenarios. In summary, PReD illustrates the potential of building a foundation model that effectively integrates domain expertise with broad generalization ability.

\begin{table*}[t]
    \centering
    \setlength{\tabcolsep}{6pt}
    \resizebox{\linewidth}{!}{%
        \begin{tabular}{lccccccc}
            \toprule
            Model                                         & MMStar        & AI2D & HallusionBench & SEEDBench IMG & POPE          & ScienceQA TEST & Average       \\
            \midrule
            \makecell[l]{GPT\mbox{-}4v \cite{OpenAI2023GPT4VSystemCard}}       & 49.7          & \textbf{75.9} & 46.5           & 71.6          & 75.4          & \textbf{82.1}           & \textbf{66.9} \\
            \makecell[l]{\textbf{PReD (Ours)} }                         & \textbf{50.2} & \textbf{74.3} & \textbf{54.5}  & \textbf{73.7} & \textbf{70.4}          & \textbf{69.2}           & \textbf{65.4}          \\
            \makecell[l]{Qwen\mbox{-}VL\mbox{-}Max \cite{Qwen-VL}}                     & 49.5          & 75.7 & 41.2           & 72.7          & 71.9          & 80.0           & 65.2          \\
            \makecell[l]{LLaVA\mbox{-}Next\mbox{-}Interleave\mbox{-}7B \cite{li2024llavanextinterleavetacklingmultiimagevideo}}  & 44.5          & 73.8 & 34.8           & 72.0          & 85.5 & 73.2 & 64.0 \\
            \makecell[l]{DeepSeek\mbox{-}VL\mbox{-}7B \cite{lu2024deepseekvl} }                  & 40.5          & 65.3 & 34.5           & 70.1          & 85.6          & 80.9           & 62.8          \\
            \makecell[l]{MiniCPM\mbox{-}V\mbox{-}2 \cite{yao2024minicpm}}                     & 39.1          & 62.9 & 36.1           & 67.1          & 86.3          & 80.7           & 62.0          \\
            \makecell[l]{ShareGPT4V\mbox{-}7B \cite{chen2023sharegpt4vimprovinglargemultimodal}}                        & 35.7          & 58.0 & 28.6           & 69.3          & \textbf{86.6}          & 69.5           & 58.0          \\
            \bottomrule
        \end{tabular}%
    }

    \caption{Results on general multimodal benchmarks.}
    \label{tab:gen_mm_bench}
\end{table*}

\section{Conclusions}

We have presented \textbf{PReD}, the first foundation model to unify the full pipeline of perception, recognition, and decision-making in the EM domain. Our core methodology involves a novel, large-scale EM corpus, PReD-1.3M, which represents signals via complementary visual views, coupled with a multi-stage curriculum to progressively instill domain expertise. Extensive experimental validations demonstrate that PReD significantly outperforms current state-of-the-art general-purpose MLLMs on PReD-Bench, a benchmark specifically tailored for evaluating EM intelligence. This substantiates PReD's superior specialized proficiency within this particular scientific domain. Crucially, PReD achieves this specialized excellence while maintaining strong, competitive performance on standard multimodal benchmarks, demonstrating that domain specialization need not come at the cost of general vision-language capabilities. Our work not only charts a scalable path for EM intelligence but also validates the immense potential of adapting vision-aligned foundation models to other complex scientific domains.

\appendix
\section*{Appendix}

This appendix provides supplementary material that delves into the architectural details, training methodology, evaluation protocols, and extensive experimental analysis of the PReD model. We begin by detailing the four-stage curriculum training pipeline for PReD (Sec.~A), covering data composition, progressive tuning strategies, and key hyperparameters. This is followed by a description of the automatic metrics used to assess free-form answers in our \texttt{OpenQA$^\#$} setting (Sec.~B). Subsequently, we present a series of complementary experiments (Sec.~C) designed to offer deeper insights, including a critical ablation study on the domain-specific expertise stage to demonstrate the prevention of catastrophic forgetting, a systematic analysis of input modality contributions, an out-of-domain generalization test, and a robustness analysis across varying Signal-to-Noise Ratios (SNR). Finally, to provide a more intuitive understanding of our work, we include additional visual and qualitative analyses (Sec.~D and beyond), featuring a hierarchical visualization of the PReD-Bench composition, a comparative radar plot of model performance, and representative case studies showcasing PReD's capabilities across the full spectrum of EM tasks.


\section{Details of Training Pipeline for PReD}

\begin{figure}[t]
    \centering
    \includegraphics[width=\linewidth]{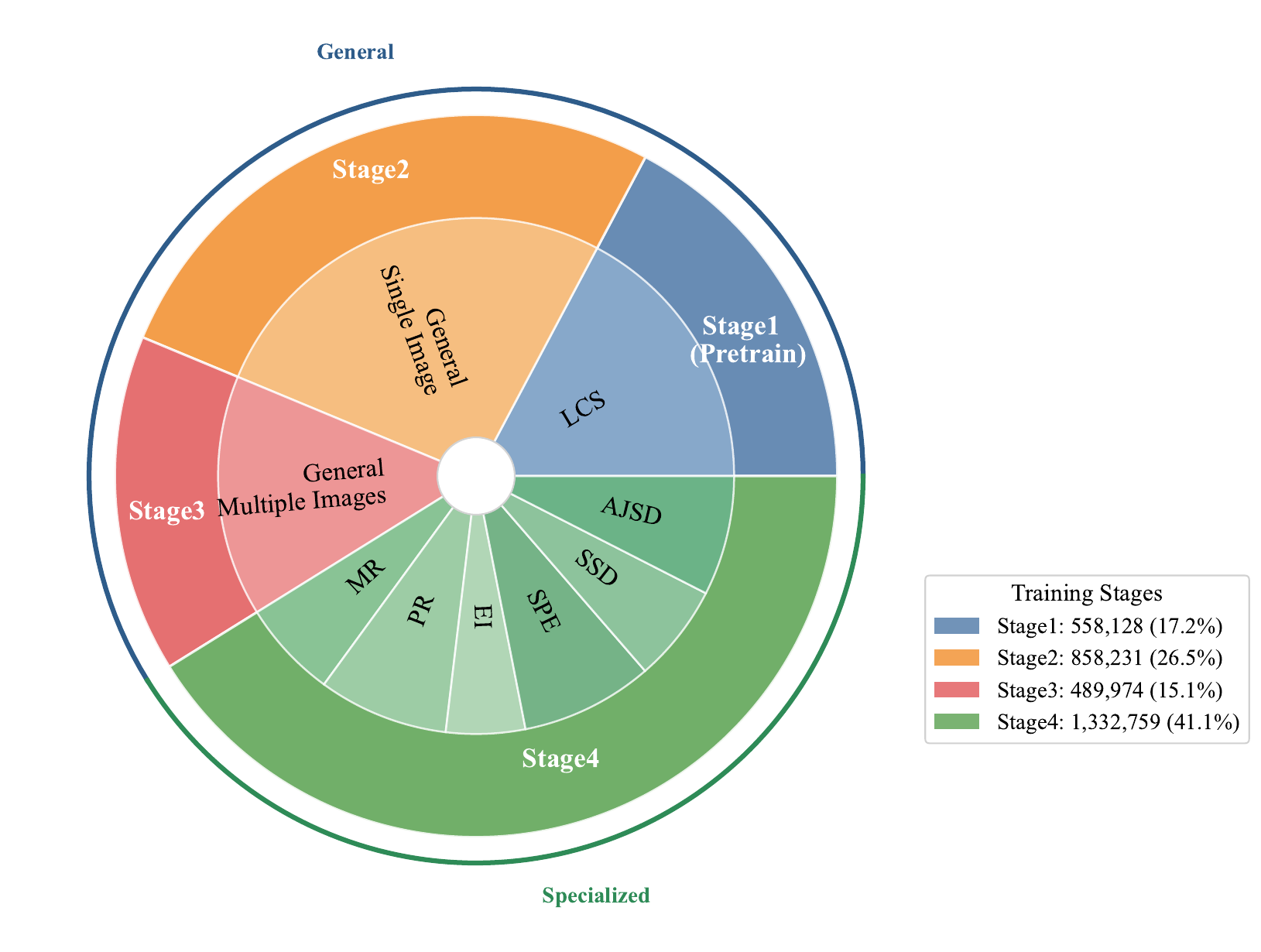}
    \caption{Overview of our four-stage curriculum and data composition. The outer ring shows the training stages; the middle ring indicates data regimes (LCS, general single/multi-image); the inner ring lists EM tasks (SSD, SSE, MR, PR, EI, AJSD) emphasized in Stage~4. Percentages denote the proportion of samples used per stage.}
    \label{fig:stage_wheel}
\end{figure}

PReD is built on Qwen-3 and a pre-trained SigLIP vision encoder, and is trained with a four-stage curriculum. Fig.~\ref{fig:stage_wheel} provides an overview of the data composition across stages, while Table~\ref{tab:hyperparams_full_stages} summarizes the corresponding hyperparameters and training schedule.

The curriculum begins with \textbf{Stage~1 (Pretrain)}, a warm-up phase that exclusively trains the \textbf{projector} using a relatively high learning rate of $1\times10^{-3}$, while both the \textbf{vision encoder (ViT)} and the \textbf{language model} remain frozen. This stage utilizes the \textbf{blip\_laion\_cc\_sbu\_558k} dataset to establish initial \textbf{vision--language alignment}. As shown in Table~\ref{tab:hyperparams_full_stages}, it operates with a base resolution of $384$, a token budget of $729$, and a \textit{flat} patch merging strategy.

From \textbf{Stage~2} onward, the full model is progressively unfrozen for \textbf{end-to-end tuning}. In \textbf{Stage~2} (26.5\%; 858,231 samples), we introduce data from \textbf{LLaVA-UHD-v2-SFT-Data} to consolidate single-image instruction-following capabilities. A \textbf{decoupled learning-rate scheme} is applied: the ViT branch is fine-tuned with $2\times10^{-6}$, while the projector and language-model parameters are updated with $1\times10^{-5}$. Correspondingly, the spatial coverage is expanded to $\mathrm{Max}\;384\times(2\times2)$ and the visual token budget is increased to $\mathrm{Max}\;729\times5$. The \textbf{patch merging strategy} is switched to \textit{spatial\_unpad} in this stage to better handle the gridded visual inputs.

\textbf{Stage~3} (15.1\%; 489,974 samples) incorporates multi-image data from the \textbf{M4-Instruct-Data} dataset to enable \textbf{multi-view modeling}. To balance computational cost and efficiency, we filter out 3D and video samples from this dataset, resulting in a final subset of 489,974 samples that focuses on static multi-view inputs. \textbf{To retain previously learned capabilities, the training set for this stage is mixed with 40\% of the general single-image corpus from Stage 2.} This stage marks a significant expansion of model capacity: the spatial coverage and token budget are further increased to their maximums of $\mathrm{Max}\;384\times(6\times6)$ and $\mathrm{Max}\;729\times10$. To accommodate the longer context, the maximum sequence length is also doubled from $4{,}096$ to $8{,}192$.

\textbf{Stage~4} (41.1\%; 1,332,759 samples) focuses on fine-tuning the model for \textbf{specialized EM tasks} using our curated \textbf{PReD-1.3M} dataset. This final stage maintains the maximal spatial coverage and token capacity established in Stage 3. \textbf{Furthermore, to prevent catastrophic forgetting of general abilities, the training corpus is augmented by mixing in 25\% of the Stage 2 single-image corpus and 50\% of the Stage 3 multi-image corpus.}

Across all four stages, optimization uses AdamW with DeepSpeed-Zero3, a cosine learning-rate schedule, weight decay of $0$, and a warmup ratio of $0.03$. The projector employs an \texttt{mlp2x2\_gelu} head, and the selected vision layer is $-2$. Each stage is trained for one epoch with per-device batch sizes of $\{8,4,2,2\}$ on a $2\times8$ A100 GPU setup.

Overall, this four-stage training pipeline starts from general visual perception (Stages~1--2), advances to multi-view processing (Stage~3), and culminates in domain-specific EM expertise (Stage~4). The combination of curriculum design, dataset mixing, and decoupled optimization balances stability and plasticity, leading to robust end-to-end EM perception, category-level recognition, and downstream decision making, while keeping the total compute and sequence lengths within the limits reported in Table~\ref{tab:hyperparams_full_stages}.

\begin{table*}[t]
    \centering
    \setlength{\tabcolsep}{6pt}
    \renewcommand{\arraystretch}{1.20}
    \resizebox{\linewidth}{!}{%
        \begin{tabular}{l|l|c|c|c|c}
            \hline
            Category & Hyperparameter                                                                & Stage-1               & Stage-2                   & Stage-3                   & Stage-4                   \\
            \hline
            \multirow{2}{*}{Vision}
            & Resolution                                                                    & $384$                 & Max $384\times(2\times2)$ & Max $384\times(6\times6)$ & Max $384\times(6\times6)$ \\
            & Tokens                                                                        & $729$                 & Max $729\times5$          & Max $729\times10$         & Max $729\times10$         \\
            \hline
            \multirow{2}{*}{Model}
            & Trainable Components                                                          & Projector             & Full Model                & Full Model                & Full Model                \\
            & Tunable Parameters                                                            & $21.5\mathrm{M}$      & $8.6\mathrm{B}$                       & $8.6\mathrm{B}$                      & $8.6\mathrm{B}$                       \\
            \hline
            \multirow{14}{*}{Training}
            & Per-device Batch Size                                                         & $8$                   & $4$                       & $2$                       & $2$                       \\
            & Gradient Accumulation                                                         & $1$                   & $1$                       & $1$                       & $1$                       \\
            & LR: $\psi_{\mathrm{ViT}}$                                                     & N/A                   & $2\times10^{-6}$          & $2\times10^{-6}$          & $2\times10^{-6}$          \\
            & LR: $\{\theta_{\mathrm{Proj}},\ \phi_{\mathrm{LLM}}\}$ & $1\times10^{-3}$ & $1\times10^{-5}$ & $1\times10^{-5}$ & $1\times10^{-5}$ \\
            & Epoch                                                                         & $1$                   & $1$                       & $1$                       & $1$                       \\
            & Optimizer                                                                     & AdamW                 & AdamW                     & AdamW                     & AdamW                     \\
            & Deepspeed                                                                     & Zero3                 & Zero3                     & Zero3                     & Zero3                     \\
            & Weight Decay                                                                  & $0$                   & $0$                       & $0$                       & $0$                       \\
            & Warmup Ratio                                                                  & $0.03$                & $0.03$                    & $0.03$                    & $0.03$                    \\
            & LR Schedule                                                                   & cosine                & cosine                    & cosine                    & cosine                    \\
            & Projector Type                                                                & \texttt{mlp2x2\_gelu} & \texttt{mlp2x2\_gelu}     & \texttt{mlp2x2\_gelu} & \texttt{mlp2x2\_gelu} \\
            & Vision Select Layer                                                           & $-2$                  & $-2$                      & $-2$                      & $-2$                      \\
            & Patch Merge Type                                                              & flat                  & spatial\_unpad            & spatial\_unpad            & spatial\_unpad            \\
            & Max Seq Length                                                                & $4{,}096$             & $4{,}096$                 & $8{,}192$                 & $8{,}192$                 \\
            & A100 GPUs                                                                      & $2\times8$            & $2\times8$                & $2\times8$                & $2\times8$                \\
            \hline
        \end{tabular}
    } 

    \caption{Training schedule across four curriculum stages.}
    \label{tab:hyperparams_full_stages}
\end{table*}

\section{Details of Evaluation Metrics}

To evaluate free-form answers in the \texttt{OpenQA$^\#$} setting, we adopt four standard automatic metrics. BLEU-4 (BLEU4) measures local n\mbox{-}gram precision (1--4 grams) with a brevity penalty~\cite{bleu02}. ROUGE is recall oriented, including ROUGE-N and ROUGE-L to assess coverage via n\mbox{-}gram recall and LCS-based F-score, respectively~\cite{rouge04}. METEOR performs alignment-based matching, combines precision and recall with a fragmentation penalty, and allows stem and synonym matches to capture semantic paraphrases~\cite{meteor05}. CIDEr employs TF--IDF-weighted n\mbox{-}gram representations and cosine similarity against multiple references to emphasize consensus with the reference set~\cite{cider15}.

\section{Complementary Experiments}

\subsection{Ablation Study on Domain-Specific EM Expertise Stage (Stage-4)}

To investigate the impact of our curriculum learning strategy, we conduct a crucial ablation study focusing on the data composition of the domain-specific EM expertise stage. We compare two models: 
1) \textbf{PReD$^-$}, which is fine-tuned in Stage 4 \emph{exclusively} on our specialized \textbf{PReD-1.3M} dataset.
2) \textbf{PReD}, which is fine-tuned on a \emph{mixture} of the PReD-1.3M dataset and the general visual corpora from the first three stages.

We evaluate both models on two fronts: their performance on specialized domain tasks and their general multimodal capabilities.

\paragraph{Performance on Domain-Specific EM Tasks.}
Our evaluation on specialized EM tasks uses two types of metrics. For the five recognition and reasoning tasks (SSD, SPE, MR, PR, EI), we report standard \textbf{accuracy}. For the text-generation task, AJSD, we calculate a composite score from four standard metrics: BLEU-4, ROUGE-L, METEOR, and CIDEr. To ensure a consistent scale for averaging across all tasks, this composite score is derived by averaging the four metrics and multiplying the result by 100.

The detailed performance comparison is presented in Table~\ref{tab:ablation_results_PReD}. The hyper-specialized PReD$^-$ model, trained without general corpora, achieves a slightly higher average score of 73.10\% compared to our final PReD model's 71.85\%. This indicates that while focusing solely on domain data yields a marginal performance gain on specialized tasks, it comes with a significant trade-off.

\paragraph{Performance on General Multimodal Benchmarks.}
The nature of this trade-off becomes clear when evaluating general capabilities, as presented in Table~\ref{tab:ablation_results}. Here, the benefit of our data mixing strategy is dramatic. Our full model, \textbf{PReD}, achieves a strong average score of 65.37\%. In stark contrast, the performance of \textbf{PReD$^-$} collapses to 35.92\%. This significant gap highlights that fine-tuning exclusively on specialized data leads to severe \textbf{catastrophic forgetting} of general multimodal knowledge acquired in earlier stages.

In summary, these results demonstrate the critical role of mixing general-domain data in the final training stage. While a narrow focus on EM data yields a slight edge (+1.25\%) in specialized accuracy, it catastrophically degrades general abilities. Our strategy successfully mitigates this issue, creating a robust model that excels in its specific domain without sacrificing its foundational multimodal competence. This balance is crucial for developing a true EM foundation model.

\begin{table*}[t]
    \centering
    \setlength{\tabcolsep}{4pt} 
    \renewcommand{\arraystretch}{1.2} 
    \resizebox{0.85\textwidth}{!}{%
        \begin{tabular}{l|ccccccc}
            \hline
            \textbf{Method} & \textbf{SSD} & \textbf{SPE} & \textbf{MR} & \textbf{PR} & \textbf{EI} & \textbf{AJSD} & \textbf{Average} \\
            \hline
            \textbf{PReD$^-$ (w/o General Corpora)} & \textbf{83.55} & \textbf{86.37} & \textbf{72.40} & \textbf{74.40} & \textbf{71.83} & \textbf{50.05} & \textbf{73.10} \\
            PReD (Ours) & 83.40 & 85.70 & 72.00 & 72.00 & 69.43 & 48.59 & 71.85 \\
            \hline
        \end{tabular}
        }
    \caption{
        \textbf{Performance comparison on the specialized EM benchmark.} 
        The variant fine-tuned exclusively on EM data (\textbf{PReD$^-$}) achieves a marginal performance gain on these domain-specific tasks. 
        As shown in Table~\ref{tab:ablation_results}, this slight advantage comes at the cost of a severe degradation in general multimodal capabilities.
    }
    \label{tab:ablation_results_PReD}
\end{table*}

\begin{table*}[t]
    \centering
    \setlength{\tabcolsep}{4pt} 
    \renewcommand{\arraystretch}{1.2} 
    \resizebox{0.99\textwidth}{!}{
        \begin{tabular}{l|c|c|c|c|c|c|c}
            \hline
            \textbf{Method} & \textbf{MMStar} & \textbf{AI2D} & \textbf{HallusionBench} & \textbf{SEEDBench IMG} & \textbf{POPE} & \textbf{ScienceQA TEST} & \textbf{Average} \\
            \hline
            PReD$^-$ (w/o General Corpora) & 39.47 & 54.44 & 8.10 & 56.75 & 0.74 & 56.02 & 35.92 \\
            \textbf{PReD (Ours)} & \textbf{50.20} & \textbf{74.32} & \textbf{54.47} & \textbf{73.67} & \textbf{70.37} & \textbf{69.16} & \textbf{65.37} \\
            \hline
        \end{tabular}}
    \caption{
        \textbf{Ablation study on mixing general-domain data in the final training stage.}
        We compare our full model (\textbf{PReD}) against a variant (\textbf{PReD$^-$}) that was fine-tuned exclusively on the specialized EM dataset without the general visual corpora.
        Mixing general-domain data significantly prevents catastrophic forgetting and boosts performance across general benchmarks.
    }
    \label{tab:ablation_results}
\end{table*}

\subsection{Ablation Study on Input Modalities}
To understand the contribution of each input modality, we conducted a comprehensive ablation study across two distinct tasks: modulation recognition (a classification task, Table~\ref{tab:modulation_combo_flags}) and anti-jamming decision-making (a generation task, Table~\ref{tab:ajsd_modality_ablation}). For all experiments, unselected modalities were replaced with zero-filled placeholders to maintain consistent input dimensionality. Our findings reveal consistent patterns across both tasks.

\paragraph{Finding 1: IQ as the Foundational Modality.}
Across both tasks, the \textbf{IQ} modality consistently proves to be the most informative when used alone. It achieves the highest single-modality accuracy in modulation recognition (50.8\%) and dominates all metrics in anti-jamming decision-making (0.363 average score). This is physically expected, as IQ data represents the lossless, raw complex-valued information of the signal containing all amplitude and phase variations. It carries the strongest discriminative cues, forming a robust foundation for downstream tasks.

\paragraph{Finding 2: Complementarity vs. Redundancy.}
The combination of modalities highlights clear synergistic and redundant relationships. For instance, in modulation recognition, \emph{Constellation+IQ} (64.2\%) significantly outperforms the sum of its parts. This demonstrates that while IQ provides temporal details, the Constellation diagram explicitly visualizes the modulation state space (symbol grid), transforming statistical distributions into distinct geometric visual patterns that are easier for the vision encoder to capture. Conversely, \emph{FFT+STFT} (37.8\%) performs poorly, indicating substantial redundancy. Since STFT already encapsulates spectral information over time, the global spectral statistics provided by FFT offer limited additional entropy compared to the time-frequency view. This pattern holds in the decision-making task, where combinations including both IQ and other modalities like \emph{STFT} or \emph{Constellation} yield top results within their cardinality groups.

\paragraph{Finding 3: The Efficiency-Accuracy Sweet Spot.}
While using all four modalities consistently yields the best performance on both tasks (72.0\% and 0.4860), a clear point of diminishing returns emerges. The three-modality combination of \emph{Constellation+STFT+IQ} achieves 70.0\% accuracy in recognition and an average score of 0.414 in decision-making. These results represent 97.2\% and 85.1\% of the full four-modality performance, respectively. This makes the \textbf{\emph{Constellation+STFT+IQ}} combination a highly effective trade-off. It successfully covers the three critical physical dimensions of EM signals: temporal dynamics (IQ), spectral footprint (STFT), and modulation topology (Constellation), rendering the marginal gain from adding the final FFT modality relatively small.

In summary, our cross-task ablation provides clear guidance for modality selection. \textbf{IQ} is indispensable as the information source. \textbf{Constellation} and \textbf{STFT} provide critical, complementary perspectives that explicitly visualize features hidden in the time domain. \textbf{FFT}, while beneficial, often overlaps with STFT and offers the least unique information. For resource-constrained applications, the \emph{Constellation+STFT+IQ} trio is an optimal choice, while the full four-modality input should be used when maximizing performance is the primary goal.

\begin{table}[t]
    \centering
    \setlength{\tabcolsep}{5pt}
    \renewcommand{\arraystretch}{1.1}
    \begin{tabular}{cccc|c}
        \hline
        \textbf{Constellation} & \textbf{FFT} & \textbf{STFT} & \textbf{IQ} & \textbf{Accuracy} \\
        \hline
        \cmark & \xmark & \xmark & \xmark & 43.4 \\
        \xmark & \cmark & \xmark & \xmark & 35.0 \\
        \xmark & \xmark & \cmark & \xmark & 30.2 \\
        \xmark & \xmark & \xmark & \cmark & \textbf{50.8} \\
        \hline
        \cmark & \cmark & \xmark & \xmark & 54.8 \\
        \cmark & \xmark & \cmark & \xmark & 58.4 \\
        \cmark & \xmark & \xmark & \cmark & \textbf{64.2} \\
        \xmark & \cmark & \cmark & \xmark & 37.8 \\
        \xmark & \cmark & \xmark & \cmark & 55.6 \\
        \xmark & \xmark & \cmark & \cmark & 57.0 \\
        \hline
        \cmark & \cmark & \cmark & \xmark & 59.2 \\
        \cmark & \cmark & \xmark & \cmark & 68.8 \\
        \cmark & \xmark & \cmark & \cmark & \textbf{70.0} \\
        \xmark & \cmark & \cmark & \cmark & 61.0 \\
        \hline
        \cmark & \cmark & \cmark & \cmark & \textbf{72.0} \\
        \hline
    \end{tabular}
    \caption{MR accuracy (\%) for different modality combinations. The optimal result is highlighted in bold.}
    \label{tab:modulation_combo_flags}
\end{table}

\begin{table*}[t]
    \centering
    \setlength{\tabcolsep}{4pt}
    \renewcommand{\arraystretch}{1.15}
    \begin{tabular}{lccccc}
        \hline
        Input Modality                  & BLEU4  & ROUGE & METEOR & CIDEr & Average   \\
        \hline
        \emph{Constellation}           & 0.069 & 0.329 & 0.227 & 0.405 & 0.258 \\
        \emph{FFT}                     & 0.056 & 0.305 & 0.194 & 0.381 & 0.234 \\
        \emph{STFT}                    & 0.094 & 0.374 & 0.267 & 0.447 & 0.2955 \\
        \emph{IQ}                      & \textbf{0.149} & \textbf{0.438} & \textbf{0.337} & \textbf{0.530} & \textbf{0.363} \\
        \hline
        \emph{Constellation+FFT}       & 0.070 & 0.331 & 0.227 & 0.409 & 0.259 \\
        \emph{Constellation+STFT}      & 0.104 & 0.384 & 0.279 & 0.465 & 0.308 \\
        \emph{Constellation+IQ}        & 0.168 & 0.464 & 0.359 & 0.549 & 0.385 \\
        \emph{FFT+STFT}                & 0.095 & 0.373 & 0.267 & 0.449 & 0.296 \\
        \emph{FFT+IQ}                  & 0.153 & 0.441 & 0.341 & 0.534 & 0.367 \\
        \emph{STFT+IQ}                 & \textbf{0.185} & \textbf{0.488} & \textbf{0.381} & \textbf{0.567} & \textbf{0.405} \\
        \hline
        \emph{Constellation+FFT+STFT}  & 0.102 & 0.379 & 0.276 & 0.464 & 0.305 \\
        \emph{Constellation+FFT+IQ}    & 0.168 & 0.463 & 0.357 & 0.548 & 0.384 \\
        \emph{Constellation+STFT+IQ}   & \textbf{0.193} & \textbf{0.499} & \textbf{0.388} & \textbf{0.575} & \textbf{0.414} \\
        \emph{FFT+STFT+IQ}             & 0.185 & 0.489 & 0.381 & 0.567 & 0.405 \\
        \hline
        \emph{Constellation+FFT+STFT+IQ} & \textbf{0.253} & \textbf{0.559} & \textbf{0.515} & \textbf{0.617} & \textbf{0.486} \\
        \hline
    \end{tabular}
    \caption{Ablation on input modalities for the anti-jamming strategy decision-making task.}
    \label{tab:ajsd_modality_ablation}
\end{table*}


\subsection{Out-of-Domain Generalization for Modulation Recognition}

To further assess the generalization capabilities of PReD beyond its training distribution, we conducted a challenging out-of-domain (OOD) evaluation. While PReD was trained for modulation recognition using only the `RadioML2016.10a` dataset, we evaluated its zero-shot performance on the unseen `RadioML2016.04c` dataset, which features different channel conditions and signal characteristics. We constructed a multiple-choice question benchmark by sampling 10 signals for each modulation type at every available Signal-to-Noise Ratio (SNR), resulting in a total of 2,200 evaluation samples.

\begin{figure}[t]
    \centering
    \includegraphics[width=\linewidth]{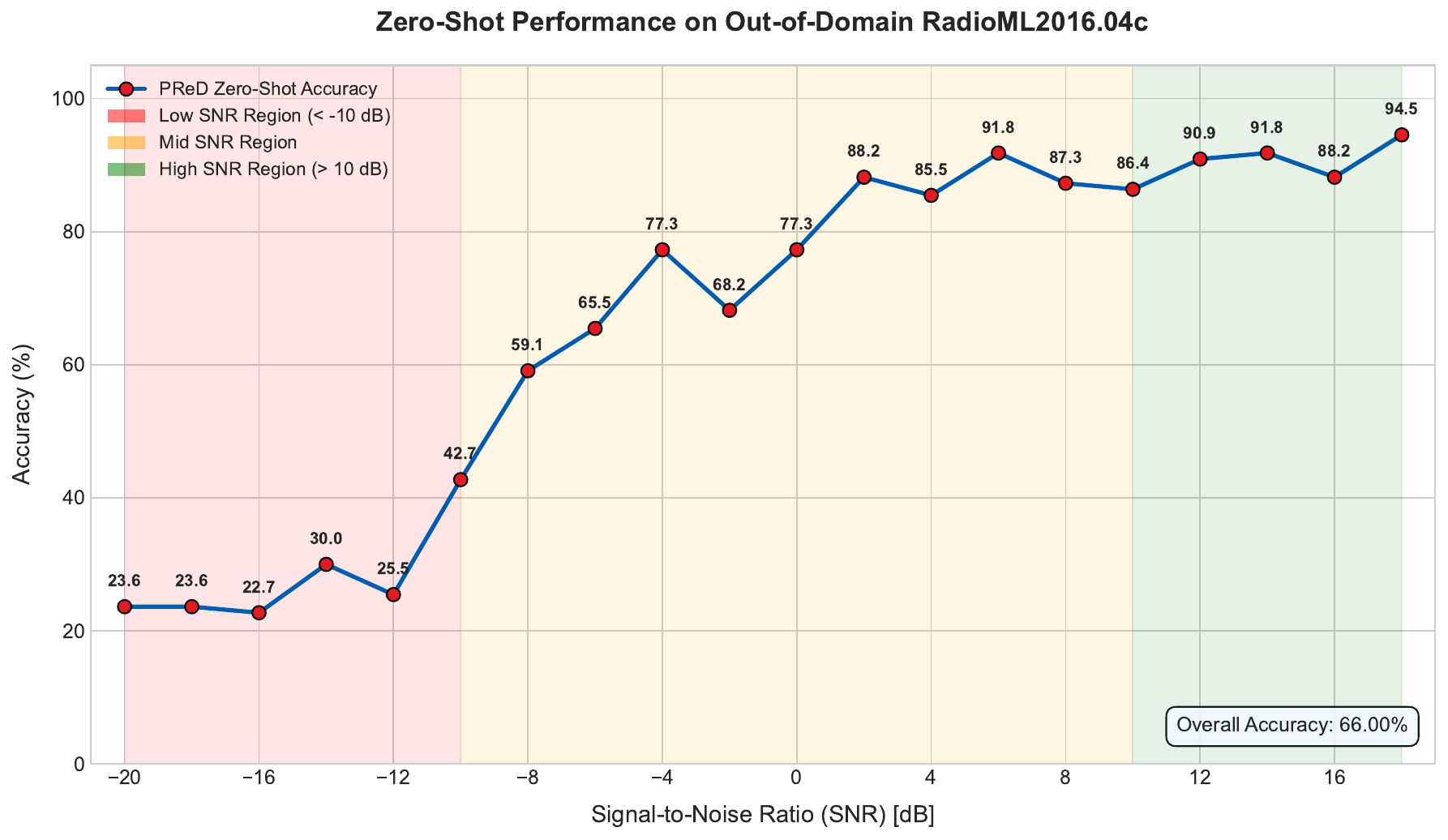} 
    \caption{
        Zero-shot modulation recognition accuracy of PReD on the OOD RadioML2016.04c dataset across SNR levels. The overall accuracy is 66.0\%.
    }
    \label{fig:snr_performance_04c}
\end{figure}

On this challenging OOD benchmark, PReD achieved a strong overall accuracy of \textbf{66.0\%} (1452/2200), demonstrating robust generalization to unseen signal environments without any fine-tuning. To gain deeper insights into the model's robustness to noise, we analyzed its performance across different SNR levels, as detailed in Figure~\ref{fig:snr_performance_04c}. The results reveal a clear and physically expected trend: performance is modest in the \textbf{low-SNR regime} ($<-10$~dB) where signals are heavily corrupted by noise; it then \textbf{improves dramatically in the mid-SNR range} ($-10$~dB to $10$~dB), and finally \textbf{stabilizes at a high level in the high-SNR range} ($>10$~dB).

This behavior, which mirrors the performance curves of specialized recognition models, confirms that PReD has learned to extract meaningful physical features from the signals rather than merely overfitting to the source dataset. This validates its potential as a robust foundation model for electromagnetic signal processing.

\subsection{Performance Analysis under Varying SNR}
To provide a more granular view of PReD's robustness, we analyze the performance of the \textbf{EM Recognition} tasks under varying SNRs. 
Within this category, we present the results for \textbf{MR} and \textbf{PR}, as their underlying benchmarks contain explicit SNR labels suitable for this analysis. 
The \textbf{EI} task is excluded from this analysis because its dataset consists of real-world captured signals that lack ground-truth SNR annotations.

\begin{figure}[t]
    \centering
    \includegraphics[width=\linewidth]{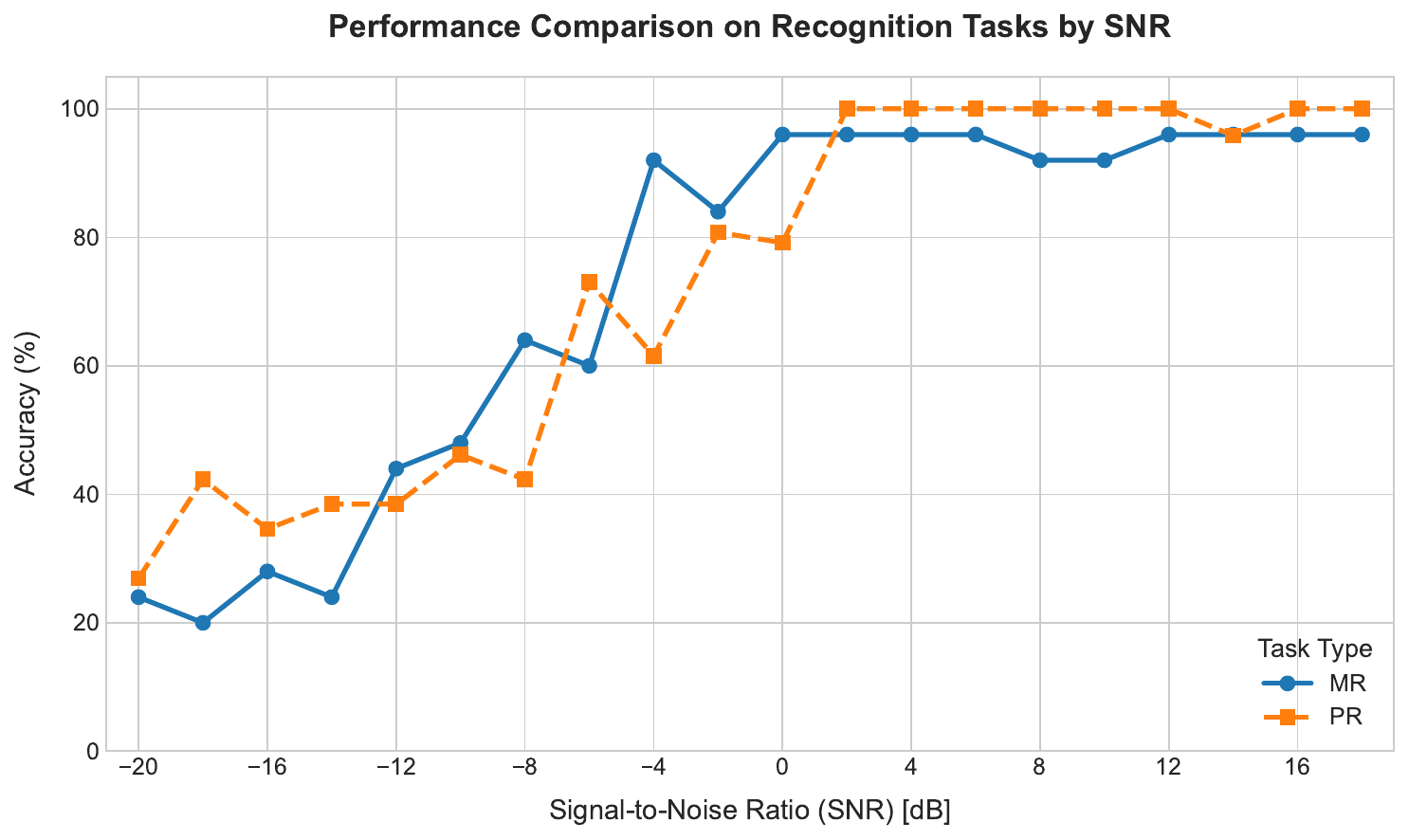} 
    \caption{
        Performance comparison on \textbf{MR} and \textbf{PR} task across varying SNR levels. Both tasks show a similar graceful degradation, highlighting the model's consistent and robust behavior against noise.
    }
    \label{fig:appendix_snr_comparison}
\end{figure}

As illustrated in Figure~\ref{fig:appendix_snr_comparison}, both tasks exhibit a remarkably consistent and physically expected trend: model accuracy degrades gracefully as the SNR decreases. This demonstrates that PReD has learned to leverage signal quality in a meaningful way, reinforcing its capability as a robust foundation model for the EM domain. The performance drop is most pronounced in the negative SNR regime, where the signal power is weaker than the noise power, yet the model retains non-trivial recognition capabilities.

\section{Visual Analysis of PReD-Bench and Benchmark Results}

This section offers a deeper, more intuitive understanding of our benchmark and results through both visual analysis and qualitative examples. First, we present two complementary visualizations: a hierarchical chart to break down the composition of PReD-Bench and a radar chart to summarize the comparative model performances discussed in the main text. Following this quantitative overview, we provide a series of representative qualitative examples to concretely illustrate PReD's capabilities across the full range of EM tasks.

\subsection{Hierarchical Composition of PReD-Bench}

\begin{figure}[t]
    \centering
    \includegraphics[width=0.8\linewidth]{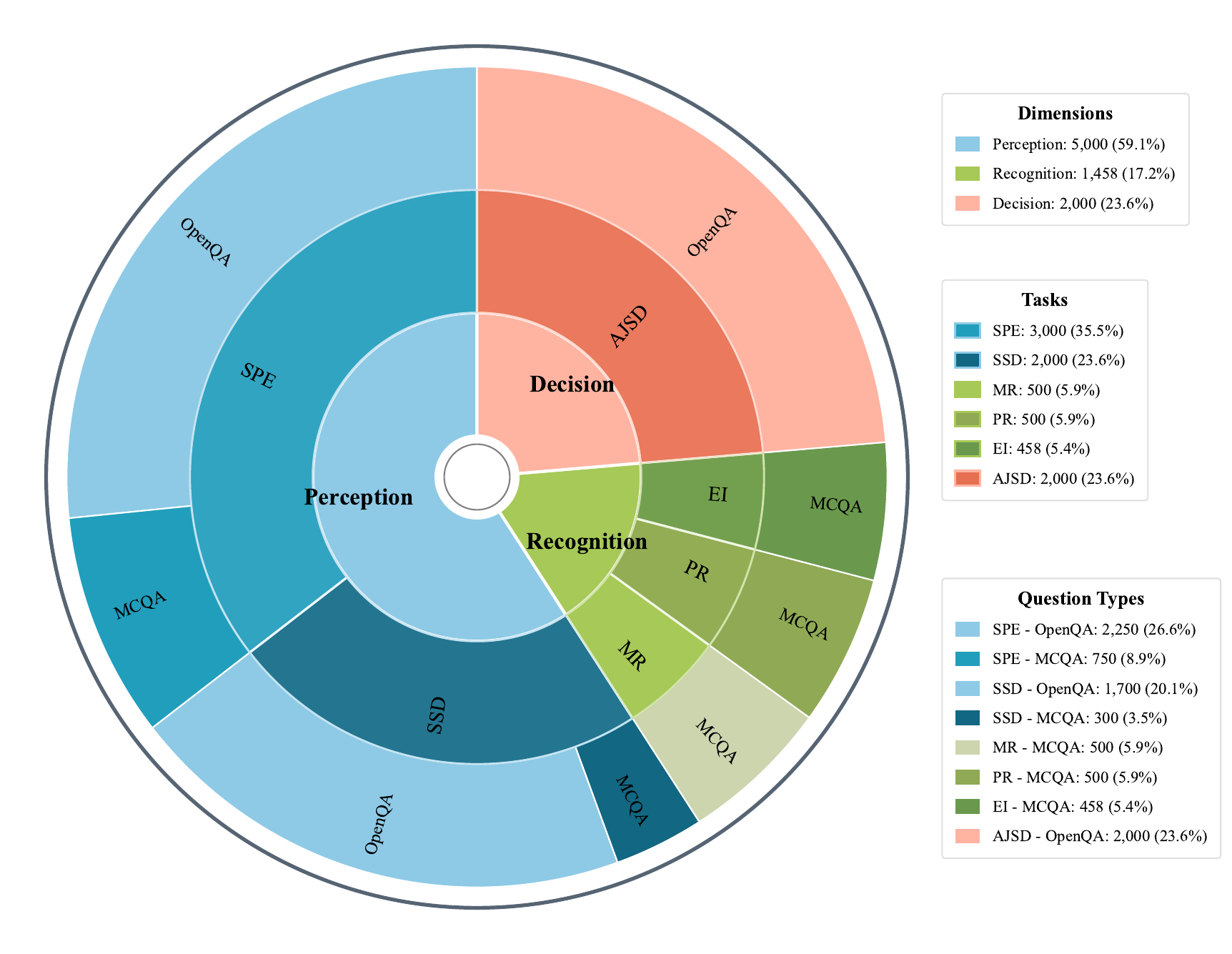} 
    \caption{Hierarchical composition of the PReD-Bench, showing the distribution across three core dimensions, six specific tasks, and two primary question types (OpenQA and MCQA).}
    \label{fig:pred-bench-sunburst}
\end{figure}

Figure~\ref{fig:pred-bench-sunburst} offers a hierarchical visualization of the PReD-Bench's composition. The chart is structured into three concentric rings representing different levels of granularity:
\begin{itemize}
    \item \textbf{Innermost Ring (Dimensions):} This ring shows the three core capabilities evaluated by the benchmark: \textbf{Perception} (59.1\%), \textbf{Recognition} (17.2\%), and \textbf{Decision-Making} (23.6\%).
    \item \textbf{Middle Ring (Tasks):} This ring breaks down the dimensions into six specific tasks. It details the sample distribution among SPE (35.5\%), SSD (23.6\%), MR (5.9\%), PR (5.9\%), EI (5.4\%), and AJSD (23.6\%).
    \item \textbf{Outermost Ring (Question Types):} The final ring specifies the question format for each task. It shows that Perception tasks (SPE, SSD) are designed with a mix of \textbf{OpenQA} and \textbf{MCQA} formats, while the majority of Recognition tasks (MR, PR, EI) are exclusively \textbf{MCQA}. The Decision-Making task (AJSD) is designed as a pure \textbf{OpenQA} generative task.
\end{itemize}
This layered structure illustrates the benchmark's comprehensive design, ensuring a balanced evaluation across diverse EM tasks and reasoning formats.

\subsection{Comparative Performance Visualization on EM Tasks}

\begin{figure}[t]
    \centering
    \includegraphics[width=0.75\linewidth]{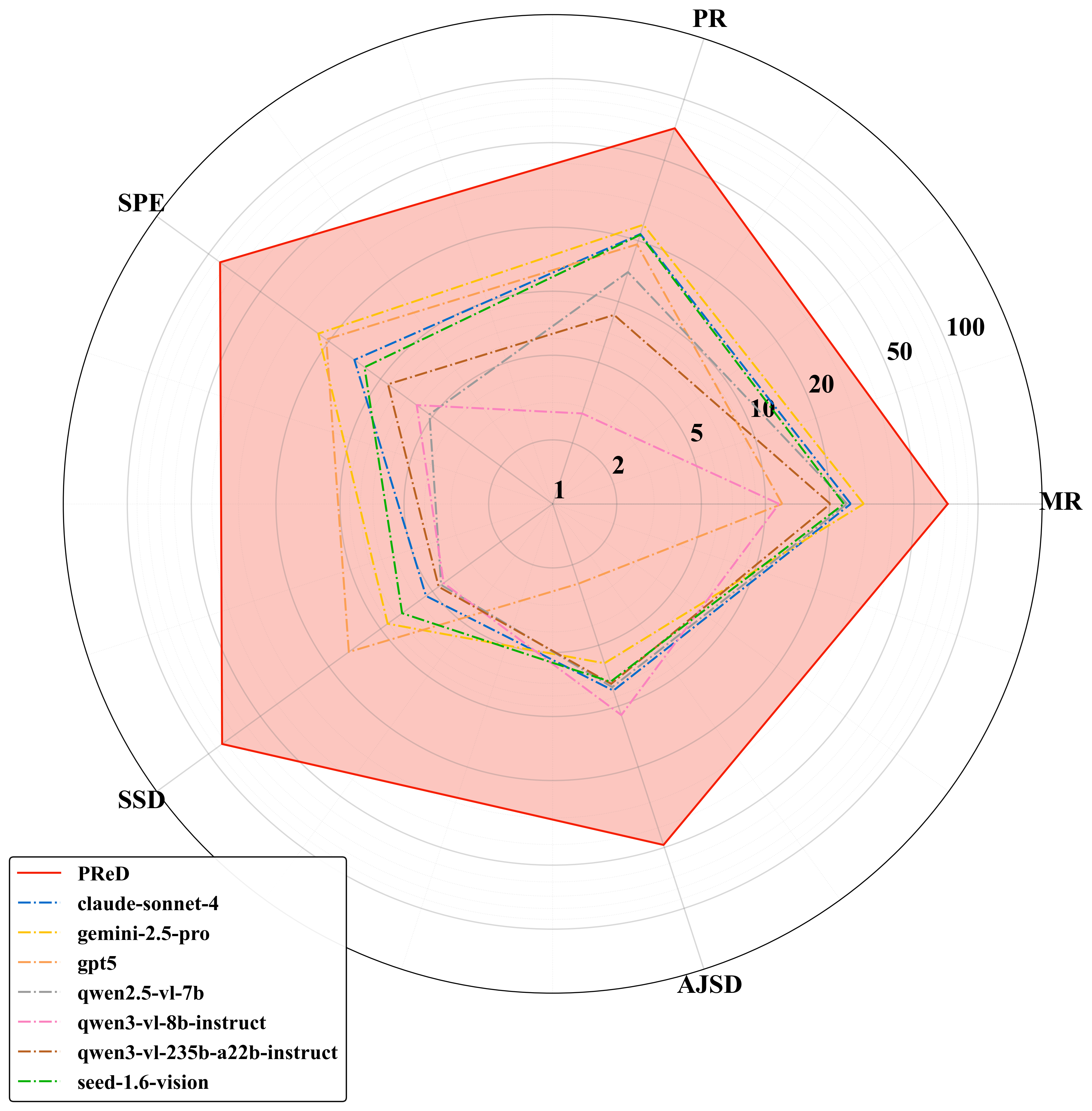} 
    \caption{Log-scale radar plot of averaged performance over five EM tasks. PReD demonstrates superior performance across all evaluated tasks.}
    \label{fig:em-radar}
\end{figure}

The log-scale radar chart in Fig.~\ref{fig:em-radar} summarizes the cross-task average performance across five representative EM tasks. \textbf{PReD consistently forms the outer envelope and yields the largest enclosed area}, highlighting its dominant advantage throughout the full EM pipeline.

In contrast, \textbf{general-purpose MLLMs exhibit a clear performance gap between foundational and advanced tasks}. They maintain moderate capabilities on foundational tasks involving \textbf{perception (SPE/SSD)} and \textbf{recognition (MR/PR)}. However, their performance collapses on the advanced, open-ended generative task of \textbf{decision-making (AJSD)}. This is particularly evident for models like GPT-5, Gemini-2.5-Pro, and Claude-Sonnet-4, all of which show a severe contraction toward the AJSD axis.

This collective pattern aligns with the tabulated results and demonstrates the critical limitation of general-purpose MLLMs when transitioning from pattern recognition to applied, open-ended reasoning. This further validates the effectiveness of our domain-specific modeling and multi-stage curriculum in building true end-to-end intelligence.

\subsection{Qualitative Examples on PReD-Bench Tasks}

\begin{figure*}[t!]
    \centering
    \setlength{\fboxsep}{1pt} 

    \begin{subfigure}[b]{0.48\textwidth} 
        \centering
        \fbox{\includegraphics[width=\linewidth, height=6cm]{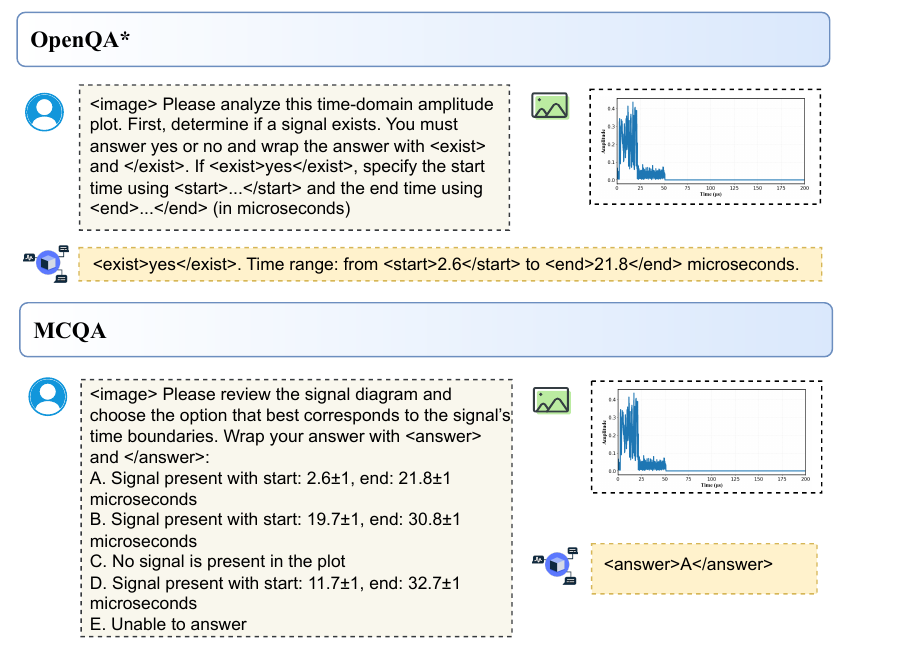}}
        \caption{Task 1: Signal Segment Detection (SSD)}
        \label{fig:example_ssd}
    \end{subfigure}
    \hfill
    \begin{subfigure}[b]{0.48\textwidth} 
        \centering
        \fbox{\includegraphics[width=\linewidth, height=6cm]{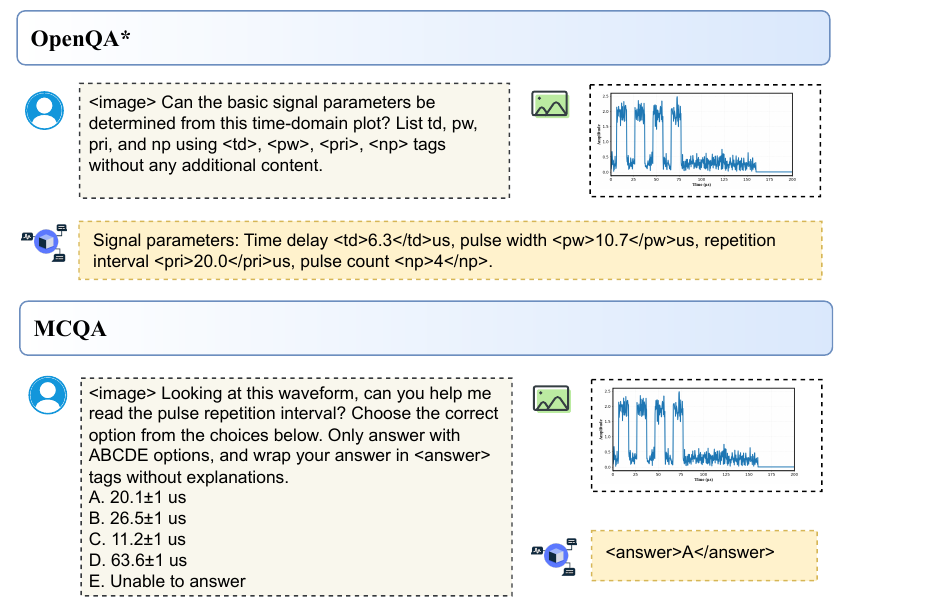}}
        \caption{Task 2: Signal Parameter Estimation (SPE)}
        \label{fig:example_spe}
    \end{subfigure}

    \vspace{0.6cm} 

    \begin{subfigure}[b]{0.48\textwidth}
        \centering
        \fbox{\includegraphics[width=\linewidth, height=6cm]{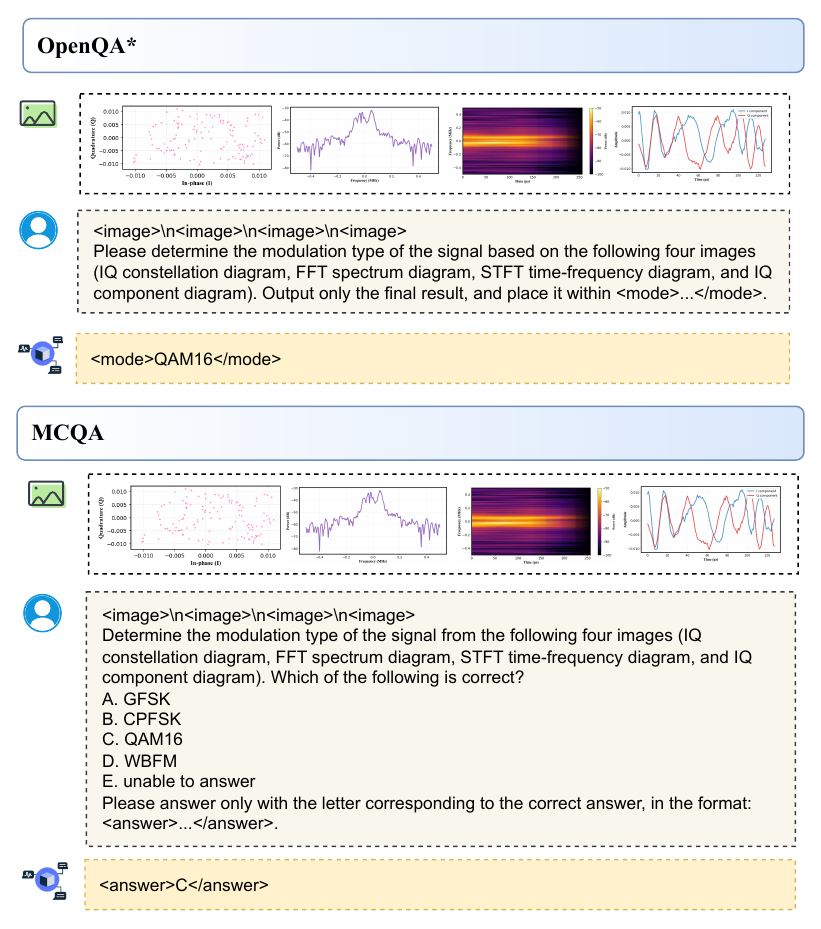}}
        \caption{Task 3: Modulation Recognition (MR)}
        \label{fig:example_mr}
    \end{subfigure}
    \hfill
    \begin{subfigure}[b]{0.48\textwidth}
        \centering
        \fbox{\includegraphics[width=\linewidth, height=6cm]{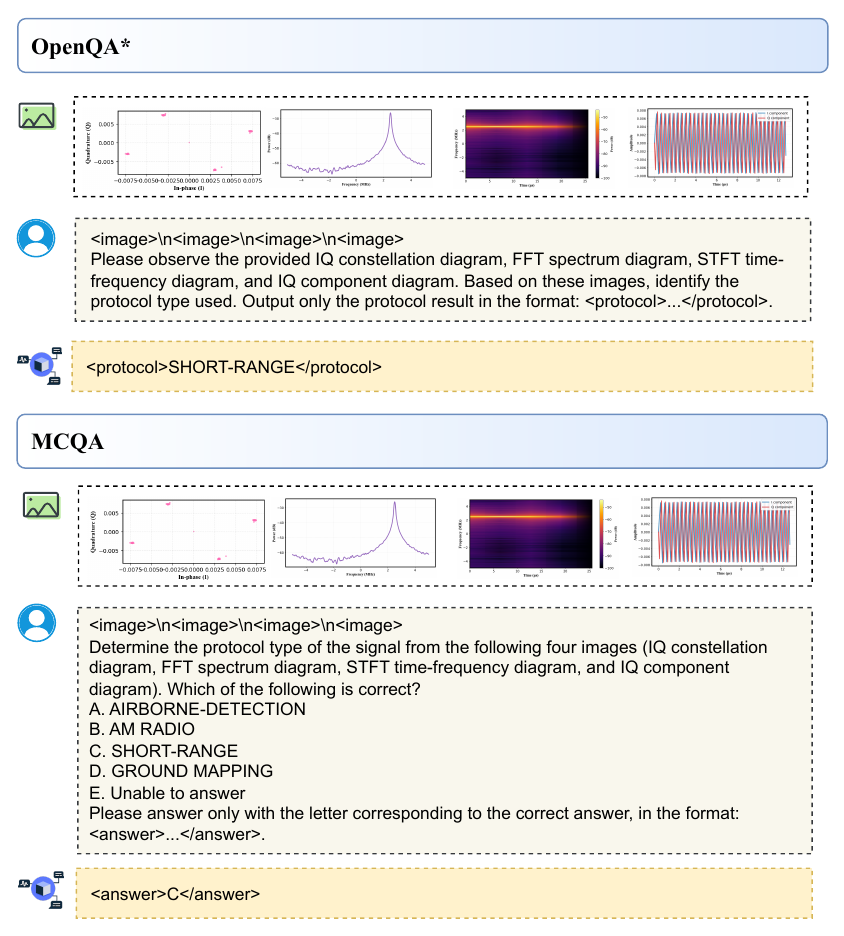}}
        \caption{Task 4: Protocol Recognition (PR)}
        \label{fig:example_pr}
    \end{subfigure}

    \vspace{0.6cm} 

    \begin{subfigure}[b]{0.48\textwidth}
        \centering
        \fbox{\includegraphics[width=\linewidth, height=6cm]{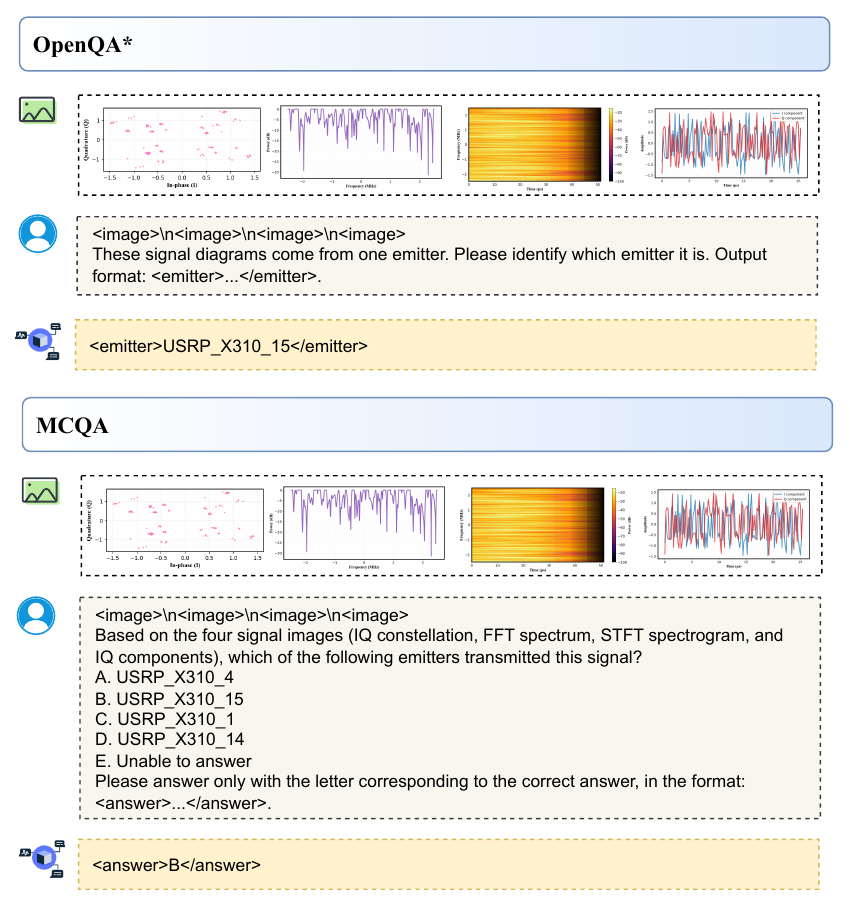}}
        \caption{Task 5: Emitter Identification (EI)}
        \label{fig:example_ei}
    \end{subfigure}
    \hfill
    \begin{subfigure}[b]{0.48\textwidth}
        \centering
        \fbox{\includegraphics[width=\linewidth, height=6cm]{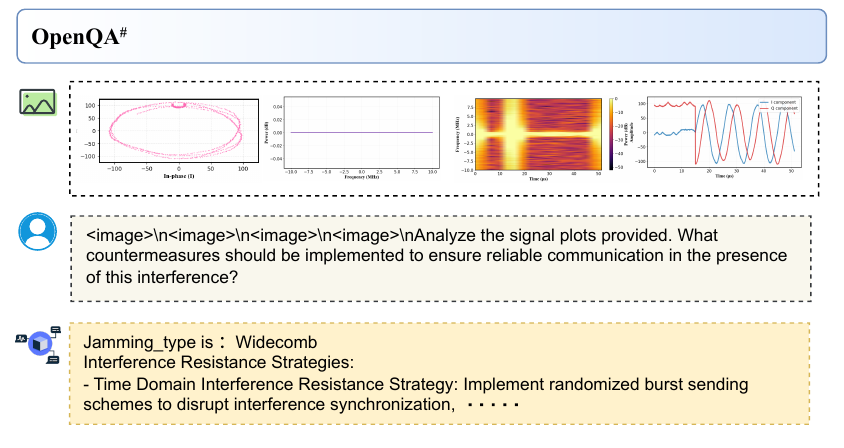}}
        \caption{Task 6: Anti-Jamming Strategy Decision-making (AJSD)}
        \label{fig:example_ajsd}
    \end{subfigure}

    \caption{
        \textbf{Qualitative examples of PReD's performance across the six core tasks of PReD-Bench.} 
        Each panel showcases a different capability, from perception (a, b), to recognition (c, d, e), and finally to complex decision-making (f).
    }
    \label{fig:qualitative_examples}
\end{figure*}

To provide a qualitative illustration of PReD's capabilities, we present representative examples for each of the six core tasks defined in the PReD-Bench. These examples, shown in Figure~\ref{fig:qualitative_examples}, showcase the model's ability to handle a diverse range of inputs and generate accurate, context-aware responses across the full spectrum of EM intelligence: from foundational perception and recognition to complex decision-making.It is worth noting that for the open-ended nature of Modulation Recognition, Protocol Recognition, and Emitter Identification tasks, where different expressions can convey the same meaning, we standardized the evaluation in our test set by removing the open-ended response format for these specific tasks to ensure consistent and objective assessment.

Each subfigure presents a user query, which may include single or multiple signal images, followed by the detailed, human-like response generated by PReD. These examples complement the quantitative results in the main paper by offering a direct view into the model's reasoning and generative abilities.

\clearpage

{
    \small
    \bibliographystyle{unsrt}
    \bibliography{main}
}


\end{document}